\documentclass{article}

\usepackage{arxiv}

\usepackage{hyperref}
\usepackage[T1]{fontenc}
\usepackage{graphicx}
\usepackage{amssymb,amsmath,amsthm}
\usepackage{mathtools}
\usepackage{multirow}
\usepackage{booktabs}	
\usepackage{float}	
\usepackage{amsmath,amsfonts,latexsym,amssymb}
\usepackage[mathscr]{euscript}
\usepackage{mathtools}
\usepackage{subfigure}
\usepackage{enumitem}
\usepackage{bm}
\usepackage{color,soul} 
\usepackage{array}
\usepackage[ruled,linesnumbered]{algorithm2e}
\usepackage{siunitx} 
\usepackage{cancel} 
\usepackage{comment} 
\SetKwComment{Comment}{$\triangleright$\ }{}

\newlength{\figwidth}
\newlength{\subfigwidth}
\newlength{\subfigheight}
\newcommand{\partialfirstderi}[2]{\ensuremath{\displaystyle{\frac {\partial #1} {\partial #2} } } }

\newcommand{\partialsecondderi}[2]{\ensuremath{\displaystyle{\frac {\partial^2 #1} {\partial {#2}^2} } } }

\newcommand{\mathand}{\ensuremath{\quad \text{and} \quad}}

\newcolumntype{d}[1]{D{.}{.}{#1}}
\newcommand{\matr}[1]{\mathbf{{#1}}}
\newcommand{\ve}[1]{\bm{{#1}}}
\newcommand{\E}[1]{\mathbb{E} \left[ #1 \right]}

\DeclareMathOperator{\R}{\mathbb{R}}

\DeclareMathOperator{\y}{\ve{y}}

\DeclareMathOperator{\tcf}{\ve{\theta}}

\DeclareMathOperator{\eye}{\matr{I}}

\DeclareMathOperator{\GP}{\mathcal{GP}}

\title{The role of surrogate models in the development of digital twins of dynamic systems}

\author{
 Souvik Chakraborty \\
  Department of Applied Mechanics,\\
  Indian Institute of Technology Delhi,\\
  Delhi, India.\\
  \texttt{csouvik41@gmail.com} \\
   \And
 Sondipon Adhikari \\
  College of Engineering,\\
  Swansea University\\
  Swansea, U.K. \\
  \texttt{s.adhikari@swansea.ac.uk} \\
  \And
 Ranjan Ganguli \\
  Department of Aerospace Engineering,\\
  Indian Institute of Science,\\
  Bangalore, India \\
  \texttt{ganguli@iisc.ac.in} \\
}
 
\begin{document}
\maketitle
\begin{abstract}
Digital twin technology has significant promise, relevance and potential of widespread applicability in various industrial sectors such as aerospace, infrastructure and automotive. However, the adoption of this technology has been slower due to the lack of clarity for specific applications. A discrete damped dynamic system is used in this paper to explore the concept of a digital twin. As digital twins are also expected to exploit data and computational methods, there is a compelling case for the use of surrogate models in this context. Motivated by this synergy, we have explored the possibility of using surrogate models within the digital twin technology. In particular, the use of Gaussian process (GP) emulator within the digital twin technology is explored. GP has the inherent capability of addressing noisy and sparse data and hence, makes a compelling case to be used within the digital twin framework. Cases involving stiffness variation and mass variation are considered, individually and jointly along with different levels of noise and sparsity in data. Our numerical simulation results clearly demonstrate that surrogate models such as GP emulators have the potential to be an effective tool for the development of digital twins. Aspects related to data quality and sampling rate are analysed. Key concepts introduced in this paper are summarised and ideas for urgent future research needs are proposed.
\end{abstract}


	
\section{Introduction}

A digital twin is a virtual model of a physical system which exists in the computer cloud. Increasingly, this physical system is called a physical twin \cite{souza2019digital}. Attempts to emulate the behaviour of physical systems are a key component of engineering and science. However, the difference between the digital twin and a computer model is that the digital twin updates itself to track the physical twin through the use of sensors, data analysis, machine learning, and the internet of things. The digital twin may also direct changes to the physical twin through signals sent to actuators on the physical twin. Ideally, physical and digital twins must achieve temporal synchronization. 

The widespread connectivity of physical systems to the internet allows the tracking of millions of physical twins in real-time. However, the challenge is to ensure that the digital twin mimics the physical system as closely as possible with the evolution of time. In other words, the digital twin must evolve with the physical twin over its life cycle. At the end of life, both digital and physical twins will terminate. The physical twins exist in the real world and the digital twin typically exists in the cloud \cite{coronado2018part}. Important problems in the development of digital twin technology are related to modelling and simulation of the physical twin, accuracy and speed of data transmission between the physical twin and the cloud where the digital twin resides, storage and processing of big data created by sensors through the life cycle, efficient computer processing of the digital twin model in the cloud, and possibly the actuation of the physical twin through signals sent by the digital twin from the cloud to actuators on the physical twin. 

Several authors have considered the concept of digital twin over the last two decades. An early paper on digital twins in the context of life prediction of aircraft structures was published by Tuegel and his co-workers \cite{tuegel}. Their objective was to use an ultrahigh fidelity model of an individual aircraft identified by its tail number to predict its structural life. This research was motivated by the growing power of computers and the evolution of high-performance computing. The key idea introduced in this paper was the need to track each individual aircraft by its tail number, which is a unique identifier. Thus, every physical aircraft in the fleet would have a separate digital twin, which will evolve differently compared to the other twins as each aircraft faces unique combinations of missions, loads, environment, pilot behaviour, sensor situation, etc. This concept of digital twins was later expanded to any physical system such as cars, computer servers, locomotives, turbines, machine tools, etc. The primary applications of digital twin have been in production, product design, and prognostics and health management \cite{tao2019digital,tao2,haag,millwater2019probabilistic,zhou2019digital}.

Digital twin technology has found many industrial applications, as summarized by Tao et al in their recent review paper \cite{tao}. They mention that digital twin is an enabling technology for realizing smart manufacturing and Industry 4.0. In Industry 4.0, factories are envisaged with wireless connectivity and sensors which allow the production line to function automatically. As physical systems become embellished with sensors, data transmission technology allows the collection of data throughout the life stage of the physical system. This data set is typically very large, and big data analytics is needed to find failure causes, streamline supply chains and enhance production efficiency.  According to this review paper, prognostics and health management is the area which has seen the maximum application of digital twin technology. Li et al \cite{li} created a digital twin based on the dynamic Bayesian network to monitor the operational state of aircraft wings. A probabilistic model was used to replace the deterministic physical model. Digital twins have also been used in prognostics and health management of cyber-physical systems and additive manufacturing processes. The fusion of the physical and the cyber systems remains the key challenge for the use of digital twins and one approach to address this problem is to make the modelling and simulation tasks more efficient. Surrogate modelling could be one approach to increase the efficiency of the simulation process. 

In \cite{Ganguli2020}, the authors proposed a physics-based digital twin model for a dynamical system. The development was based on the assumption that the properties of the physical system typically evolve much more slowly than real-time. This evolution thus takes place in slow time as compared to real-time. A single degree of freedom system model was created to explain the concept of digital twin and study the effect of changes in system mass and stiffness properties. Closed-form solutions to obtain the evolution of mass and stiffness with the slow time was proposed. In case, the data collected is clean, the proposed closed-form solutions yield exact estimates of the mass and stiffness evolution. However, the proposed physics-based digital twin has two major limitations. First, the physics-based digital twin proposed only yields the mass and stiffness at the slow time-step, i.e., the time-steps at which we have sensor measurements. In other words, the physics-based digital twin is not capable of providing the mass and stiffness at the intermediate and future time. Second, the physics-based digital twin can fail if the data collected is contaminated by noise. This is a major limitation as in real-life, data collected is always corrupted by noise.

From this discussion, it is clear that there is a lack of clarity and specificity regarding digital twins. Although there exists a broad consensus about what a digital twin is, there is almost no detailed methodology on how to develop one for a given system. In this work, we introduce the concept of surrogate models within the digital twin framework. This is motivated by the fact that a physics-driven digital twin, such as the one proposed in \cite{Ganguli2020, Guivarch2019} can only provide estimates at certain discrete time-steps (the time-steps corresponding to sensor measurements). Moreover, such digital twins are likely to yield erroneous results when the data collected is corrupted by complex noise. By definition, a surrogate model can be considered to be a {\it proxy} to the actual high-fidelity model. Popular surrogate models available in the literature include analysis-of-variance decomposition \cite{Chakraborty2016polynomial}, polynomial chaos expansion \cite{Sudret2008global,Xiu2002the}, support vector machines \cite{Zhao2016a,Gunn1997support,ISI:000298711400003}, Neural networks \cite{goswami2020transfer,chakraborty2020simulation,chakraborty2020transfer} and Gaussian process (GP) \cite{Bilionis2013multi,Bilionis2012multi,Chakraborty2019graph}. Successful use of surrogate models can be found in various domains including, but not limited to, stochastic mechanics \cite{Wu2016a,jp93}, reliability analysis \cite{Chakraborty2017hybrid, Dubourg2011adaptive} and optimisation \cite{Goswami2019threshold, Rahman2010reliability, kapteyn2019distributionally}. The objective of this paper is to
illustrate how surrogate models can be integrated with the digital twin technology and what possible advantages can be achieved from this integration.

Although all the surrogate models discussed above can be used within the digital twin framework, we explore the use of GP. In essence, GP is a probabilistic machine learning technique that tries to infer a distribution over functions and then utilise the same to make predictions at some unknown points. The advantage of GP over other surrogate models is two-fold. Firstly, GP being a probabilistic surrogate model is immune to over-fitting. This is extremely important for the digital twin as the data collected are corrupted by complex noise and any form of over-fitting will result in erroneous predictions. Secondly, GP is able to quantify the uncertainty due to limited and noisy data. This, in turn, can be used in the decision-making process using a digital twin. In this paper, we have illustrated the use of {\it vanilla} GP as a surrogate with the digital twin model.

The paper is organized as follows. In \autoref{sec:dt_intro}, the equation of motion of an SDOF digital twin is introduced using multiple time-scales. The fundamentals of GP are discussed in \autoref{sec:gp}. The development of a digital twin using only the mass evolution is considered in \autoref{sec:gp_DT}. Three separate cases with (a) only mass evolution, (b) only stiffness evolution and (c) both mass and stiffness evolution are considered. Numerical examples are given to illustrate the proposed ideas. Some critical discussion on the proposed methodology as well as the overall development of digital twin is given in \autoref{sec:discussions} and concluding remarks are given in \autoref{sec:conclusions}.

\section{The dynamic model of the digital twin}
\label{sec:dt_intro}

A single degree of freedom dynamic system is considered to explore the concept of a digital twin. A key idea is that a digital twin starts its journey from a `nominal model'.  The nominal model is therefore, the `initial model' or the `starting model' of a digital twin. For engineering dynamic systems, the nominal model is often a physics based model which has been verified, validated and calibrated. For example, this can be a finite element model of a car or an aircraft when the product leaves the manufacturer facility and is ready to go into service. Another key characteristics of a digital twin is its connectivity. The recent development of the Internet of Things brings forward numerous new data technologies and consequently drives the development of digital twin technology. This enables connectivity between the physical SDOF system and its digital counterpart. The basis of digital twins is based on this connection, without it, digital twin technology cannot exist. This connectivity is created by sensors on the physical system which obtain data and integrate and communicate this data through various integration technologies. 

\begin{figure}[ht!]
~\\
    \centering
    ~\\
    \includegraphics[width=\figwidth]{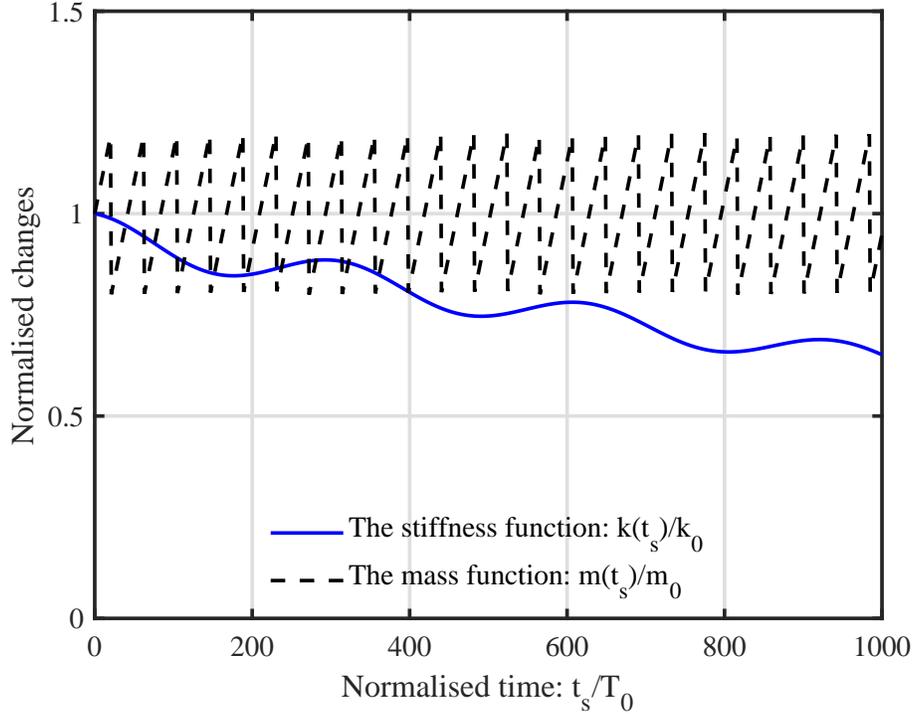}
    \caption{\label{fig:property_changes} Examples of model functions representing long term variabilities in the mass and stiffness properties of a digital twin system.}
\end{figure}

We consider a physical system which can be well approximated by a single degree of freedom spring, mass and damper system as before \cite{Ganguli2020}. It is reasonable to assume that the sensors sample data intermittently, typically, $t_s$ represents discrete time points. It is assumed that changes in $k(t_s)$, $m(t_s)$ and $c(t_s)$ as so slow that the dynamics of the system is effectively decoupled from these functional variations. The equation of motion can be written as
\begin{equation}\label{eq:sdof_dt}
m(t_s) \partialsecondderi{u(t,t_s)}{t} + c(t_s) \partialfirstderi{u(t,t_s)}{t} + k(t_s) u(t,t_s)=f(t,t_s)
\end{equation}
Here $t$ and $t_s$ are the system time and a ``slow time", respectively. $u(t,t_s)$ is a function of two variables and therefore the equation of motion is expressed in terms of the partial derivative with respect to the time variable $t$. The slow time or the service time $t_s$ can be considered as a time variable which is much slower than $t$.For example, it could represent the number of cycles in an aircraft. Thus, mass $m(t_s)$, damping $c(t_s)$, stiffness $k(t_s)$ and forcing $F(t, t_s)$ change with $t_s$, for example due to system degradation during its service life. The forcing is also a function of time $t$ and slow time $t_s$, as is the system response $x(t,t_s)$. \autoref{eq:sdof_dt} is considered as a digital twin of a SDOF dynamic system. When $t_s=0$, that is at the beginning of the service life of the system, the digital twin \autoref{eq:sdof_dt} reduces to the nominal system
\begin{equation}\label{eq:nominal_sys}
    m_0\frac{\text d^2u_0(t)}{\text dt} + c_0\frac{\text d u_0(t)}{\text dt} + k_0 u_0(t) = f_0(t),
\end{equation}
where $m_0$, $c_0$, $k_0$ and $f_0$ are respectively the mass, damping, stiffness and force at $t=0$. It is assumed that sensors are deployed on the physical system and take measurements at locations of time defined by $t_s$. The functional form of the relationship of mass, stiffness and forcing with $t_s$ is unknown and needs to be estimated from measured sensor data. We assume that damping is small so that the effect of variations in $c(t_s)$ is negligible. In effect, only variations in the mass and stiffness are considered. Without any loss of generality, the following functional forms are considered 
\begin{equation}\label{cond2}
\begin{split}
& k(t_s) = k_0(1+\Delta_k(t_s)) \\
\mathand 
& m(t_s) = m_0(1+\Delta_m(t_s))
\end{split}
\end{equation}  
In general $k(t_s)$ is expected to be a decaying function over a long time to represent a loss in the stiffness of the system. On the other hand, $m(t_s)$ can be an increasing or a decreasing function. For example, in the context of aircraft, it can represent the loading of cargo and passengers and also represent the use of fuel as the flight progresses. The following representative functions have been chosen as examples   
\begin{align}
\label{var_fnK}
&\Delta_k(t_s) = e^{-\alpha_k t_s} {\frac{(1+ \epsilon_k \cos(\beta_k t_s))} {(1+\epsilon_k)} }-1 \\
\label{var_fnm}
\mathand 
&\Delta_m(t_s) = \epsilon_m \text{ SawTooth} (\beta_m ( t_s - \pi/\beta_m) )
\end{align} 
Here SawTooth$(\bullet)$ represents a sawtooth wave with a period $2\pi$. In \autoref{fig:property_changes} overall variations in stiffness and mass properties arising from these function models have been plotted as function of time normalised to the natural time period of the nominal model.
Numerical values used for these examples are $\alpha_k=4\times 10^{-4}$,  $\epsilon_k=0.05$,  $\beta_k=2\times 10^{-2}$,  $\beta_m=0.15$ and  $\epsilon_m=0.25$. The choice of these functions is 
motivated by the fact that the stiffness degrades over time in a periodic manner representing a possible fatigue crack growth in an aircraft over repeated pressurisation. On the other hand, the mass increases and decreases over the nominal value due to re-fuelling and fuel burn over a flight period. The key consideration is that a digital twin of the dynamical system should  track these types of changes by exploiting sensor data measured on the system.

\section{Overview of Gaussian process emulators}
\label{sec:gp}

One of the most popular machine learning techniques is the Gaussian process (GP) \cite{Bilionis2013multi,Bilionis2012multi}. Gaussian process has been successfully used in the context of structural dynamic analysis  \cite{jp69} and finite element method \cite{jp137}. Unlike frequentist machine learning techniques such as the analysis-of-variance decomposition \cite{Chakraborty2016polynomial}, support vector machine \cite{Gunn1997support} and neural networks \cite{Dai2015a}, GP adopts an optimal approach and tries to infer a \textit{distribution over functions} and then utilise the same to make predictions at some unknown points \cite{Murphy2012}. We consider, $\ve{\ell} \in \R^d$ as the input variable and 
\begin{equation}\label{eq:resp}
    y = g(\ve{\ell}) + \nu
\end{equation}
to be a set of noisy measurements of the response variable, where $\nu$ represents the noise. With this setup, the objective is to estimate the latent (unobserved) function $g(\ve{\ell})$ that will enable prediction of the response variable, $\hat{y}$ at new values of $\ve{\ell}$. In GP based regression, we define a GP defined over $g(\ve{\ell})$ with the mean $\mu(\ve{\ell})$ and covariance function $\kappa(\ve{\ell},\ve{\ell}';\tcf)$	
\begin{equation} \label{eq:gpdefn}
	g(\ve{\ell})  \sim \GP \left(\mu(\ve{\ell}), \kappa(\ve{\ell},\ve{\ell}';\tcf) 
	\right),
	\end{equation}
In the above equation
	\begin{align}
	\begin{split}
	\mu(\ve{\ell}) &= \E{ g(\ve{\ell}) },\\
	\kappa(\ve{\ell},\ve{\ell}';\tcf) &= \E{ \left(g(\ve{\ell}) - 
	\mu(\ve{\ell})\right) 
	\left(g(\ve{\ell}') - \mu(\ve{\ell}')\right) }.
	\end{split}
	\end{align}
and $\tcf$ denotes the hyperparameters of the covariance function $\kappa$. The choice of the covariance function $\kappa$ allows encoding of any prior knowledge about $g(\ve{\ell})$ (e.g., periodicity, linearity, smoothness), and can accommodate approximation of arbitrarily complex functions \cite{williams2006gaussian}. The notation in \autoref{eq:gpdefn} implies that any finite collection of function values has a joint multivariate Gaussian distribution, that is $\left(g(\ve{\ell}_1), g(\ve{\ell}_2), \ldots , g(\ve{\ell}_N)\right) \sim \mathcal{N}(\ve{\mu}, \matr{K})$, where $\ve{\mu} = 	\left[\mu({\ve{\ell}_1}),\ldots,\mu({\ve{\ell}_N})  \right]^T$ is the mean vector and $\matr{K}$ is the covariance matrix with $\matr{K}(i,j) = \kappa(\ve{\ell}_i, \ve{\ell}_j)$ for $i, j = 1, 2, \ldots, N$. If no prior information is available about the mean function, it is generally set to zero, i.e.\ $\mu(\ve{\ell}) = 0$. However, for the covariance function, any function $\kappa(\ve{\ell}, \ve{\ell}')$ that generates a positive, semi-definite, covariance matrix $\matr{K}$ is a valid covariance function. With this set-up, the objective of GP is to estimate the hyperparameters, $\bm \theta$, based on the observed input-output pairs, $\left\{\ve{\ell}_j, y_j\right\}_{j=1}^{N_t}$, where $N_t$ is the number of training samples. In general, this is achieved by maximising the likelihood of the data \cite{Chakraborty2019graph}. Alternatively, one can choose to adopt a Bayesian approach to compute the posterior of the hyperparameters, $\bm \theta$ \cite{Bilionis2013multi}. Once $\bm \theta$ have been computed, the predictive distribution of $g({\ell}^*)$ given the dataset $\ve{\ell}, \y$, hyperparameters $\tcf$ and new inputs, ${\ell}^*$ is represented as
\begin{equation}\label{eq:pred_dist_gp}
    p(g({\ell}^*)|\y, \ve{\ell}, \tcf, {\ell}^*) = \mathcal{N} \left(g({\ell}^*) \; | \; 
	\mu_{\GP}({\ell}^*),  
	\sigma^2_{\GP}({\ell}^*)  \right),
\end{equation}
where 
\begin{align}
	\begin{split}
	\mu_{\GP}({\ell}^*) &= \ve{k}^T(\ve{\ell}, {\ell}^*; \tcf) 
	\left[\matr{K}(\ve{\ell},\ve{\ell}; \tcf) + \sigma^2_n \eye\right]^{-1} 
	\y,\\
	\sigma^2_{\GP}(\ell^*) &= k(\ell^*,\ell^*; \tcf) -  \ve{k}^T(\ve{\ell}, 
	\ell^*; \tcf) 
	\left[\matr{K}(\ve{\ell},\ve{\ell}; \tcf) + \sigma^2_n \eye\right]^{-1} 
	\ve{k}(\ve{\ell}, 
	\ell^*; \tcf).
	\end{split}
	\end{align}
For further details on GP, interested readers may refer \cite{rasmussen2010gaussian}.
Next, we employ GP to obtain digital twin of a single degree of freedom damped dynamic systems.  

\section{Gaussian process based digital twin}
\label{sec:gp_DT}

The development of digital twin concept hinges on the advances in the sensor technology. Using modern sensors, it is possible to collect different types of data such as acceleration time-history, displacement time-history \textit{etc}. In this paper, we assume that using sensors, the natural frequency of the system described in \autoref{eq:sdof_dt} have been measured at some distinct time-steps. Wireless sensor technology is making such measurements and transmission possible \cite{sony2019literature}. For example, Wang {\it et al.} \cite{wang2016dynamic} showed that the natural frequency of a system can be measured by using sensor technology, a fact that we use in this paper.

\subsection{Digital twin via stiffness evolution}\label{subsec: dt_stiff}
\subsubsection{Formulation}
In the first case, we assume that the change in the natural frequency with time is due to the deterioration of the stiffness of the system only. With this, the equation of motion of the digital twin is represented as
\begin{equation}\label{eq:dt_stff_only_ode}
    m_0\frac{\text d^2u\left( t \right)}{\text dt^2} + c_0\frac{\text du\left( t \right)}{\text dt} + k\left(t_s\right)u\left( t \right) = f\left(t\right),
\end{equation}
where 
\begin{equation}\label{eq:stiff_evol}
    k\left(t_s\right) = k_0\left(1 + \Delta_k\left(t_s\right)\right).
\end{equation}
\autoref{eq:dt_stff_only_ode} is a special case of \autoref{eq:sdof_dt}.
Note that in a practical scenario, we generally have no prior information about how the stiffness will deteriorate.
We postulate that the deterioration of stiffness can be represented by using a GP
\begin{equation}\label{eq:gp_gt_stiff}
    \Delta_k\left(t_s\right) \approx \Delta_{\hat k} \left(t_s\right) \sim \mathcal{GP}\left(\mu_k\left(t_s\right), \kappa_k \left(t_s, t_s';\bm \theta \right) \right).
\end{equation}
However, at this stage, we have no measurements corresponding to $\Delta_{\hat k}\left(t_s\right)$ and hence, it is not possible to estimate the hyperparameters, $\bm \theta$ of the GP. We substitute Eqs.~(\ref{eq:stiff_evol}) and (\ref{eq:gp_gt_stiff}) into \autoref{eq:dt_stff_only_ode}. With this, the governing differential equation of the digital twin can be represented as
\begin{equation}\label{eq:gp_dt_stff_only_ode}
    m_0\frac{\text d^2u\left( t \right)}{\text dt^2} + c_0\frac{\text du\left( t \right)}{\text dt} + k_0\left(1 + \Delta_{\hat k}\left(t_s\right)\right)u\left( t \right) = f\left(t\right).
\end{equation}
Solving \autoref{eq:gp_dt_stff_only_ode}, the natural frequency of the system is obtained as
\begin{equation}\label{eq:freq_dt}
    \lambda_{s_{1,2}} \left(t_s\right) = -\zeta_0\omega_0 \pm \text i \omega_0\sqrt{1 + \Delta_{\hat k}\left(t_s\right) - \zeta_0^2},
\end{equation}
where $\omega_0$ and $\zeta_0$ are the natural frequency and damping ratio of the system at $t_s = 0$.
Rearranging \autoref{eq:freq_dt}, we have
\begin{equation}
    \lambda_{s_{1,2}} \left(t_s\right) = - \underbrace{\frac{\zeta_0}{\sqrt{1 + \Delta_{\hat k}\left(t_s\right)}}}_{\zeta_s\left(t_s\right)} \underbrace{\omega_0 \sqrt{1 + \Delta_{\hat k}\left(t_s\right)}}_{\omega_s\left(t_s\right)} \pm \text{i} \underbrace{\omega_0\sqrt{1 + \Delta_{\hat k}\left(t_s\right)}\sqrt{1 - \left(\frac{\zeta_0}{\sqrt{1 + \Delta_{\hat k} \left( t_s \right)}}\right)^2}}_{\omega_{d_s}\left(t_s\right)},
\end{equation}
where 
$\omega_s\left(t_s\right) = \omega_0 \sqrt{1 + \Delta_{\hat k}\left(t_s\right)}$ is the evolution of the natural frequency, 
$\zeta\left(t_s\right) =\zeta_0 /\sqrt{1 + \Delta_{\hat k}\left(t_s\right)}$ is the evolution of the damping factor
and 
$\omega_{d_s}\left(t_s\right) = \omega_s \left(t_s\right) \sqrt{1 - \zeta_s^2 \left( t_s \right)}$ is the evolution of the damped natural frequency with slower time-scale $t_s$. Since most of the natural frequency extraction techniques extract the damped natural frequency of the system, we assume that the data obtained using the sensor to be the same. Following Ganguli and Adhikari \cite{Ganguli2020}, it can be shown that
\begin{equation}\label{eq:delta_k}
    \Delta_{\hat k}\left(t_s \right) = - \tilde d_1 \left(t_s\right)\left(2 \sqrt{1 - \zeta_0^2} - \tilde d_1\left(t_s \right) \right),
\end{equation}
where
\begin{equation}\label{eq:abs_dist}
    \tilde d_1 \left(t_s \right) = \frac{d_1\left(\omega_{d_0}, \omega_{d_s}\left(t_s\right) \right)}{\omega_0}.
\end{equation}
The function $d_1\left( \omega_{d_0}, \omega_{d_s}\left(t_s\right) \right)$ in \autoref{eq:abs_dist} is the 
distance between $\omega_{d_0}$ and $\omega_{d_s}\left(t_s\right)$
\begin{equation}
    d_1\left( \omega_{d_0}, \omega_{d_s}\left(t_s\right) \right) = \left||\omega_{d_0} -  \omega_{d_s}\left(t_s\right) \right||_2.
\end{equation}
Now given the fact that the initial damped frequency of the system, $\omega_{d_0}$ is known and we have sensor measurements for $\omega_{d_s}\left(t_s\right)$, one can easily compute $\tilde d_1 \left(t_s \right)$ and $\Delta_{\hat k} \left(t_s\right)$ by using \autoref{eq:abs_dist} and substituting it into \autoref{eq:delta_k}. Having said that, it is only fair to assume that the sensor measurements, $\omega_{d_s}\left(t_s\right)$, will be corrupted by some noise and hence, the estimates of $\Delta_{\hat k} \left(t_s\right)$ are also noisy. Nonetheless, by considering these noisy estimates of $\Delta_{\hat k} \left(t_s\right)$ to be training outputs corresponding to the inputs $t_s$, we obtain the hyperparameters ($\bm \theta$) of the GP described in \autoref{eq:gp_gt_stiff}. In this work, we use Bayesian optimisation \cite{pelikan1999boa} to maximise the likelihood of the data. 
Within the Bayesian optimization framework, L-BFGS optimizer is used with a tolerance of $10^-5$. 
To avoid local convergence, multiple starting points were used.
The hyperparameter $\bm \theta$ completely describes the digital-twin in \autoref{eq:gp_dt_stff_only_ode}. One important question associated with GP is the form of the covariance function and the mean function. In this work, we have considered a pool for the mean function and the covariance function and then selected the most suitable mean function and covariance function based on Bayesian information criteria. The possible candidates for the mean function and the covariance function are shown in \autoref{tab:mean_cov_pool}. Functional form of these covariance kernels can be found in \cite{williams2006gaussian}. Details on Bayesian information criteria are presented in Appendix \ref{app}.
\begin{table}[ht!]
    \caption{Candidate functions for mean and covariance functions in GP}
    \label{tab:mean_cov_pool}
    \centering
    \begin{tabular}{p{4 cm}p{10 cm}}
    \hline
        \textbf{Functions} & \textbf{Candidates} \\ \hline
        Mean function & Constant, Linear, Quadratic \\ 
        & \\
         {Covariance function} & Exponential, Squared Exponential, Matern 3/2, Matern 5/2, Rational Quadratic, ARD Exponential, ARD Squared Exponential, ARD Matern 3/2, ARD Matern 5/2, ARD Rational Quadratic \\ \hline
    \end{tabular}
\end{table}

\subsubsection{Numerical illustration}

\begin{figure}[ht!]
  \centering
  \subfigure[Changes in the damped natural frequency with time]{
  \includegraphics[width=0.48\textwidth]{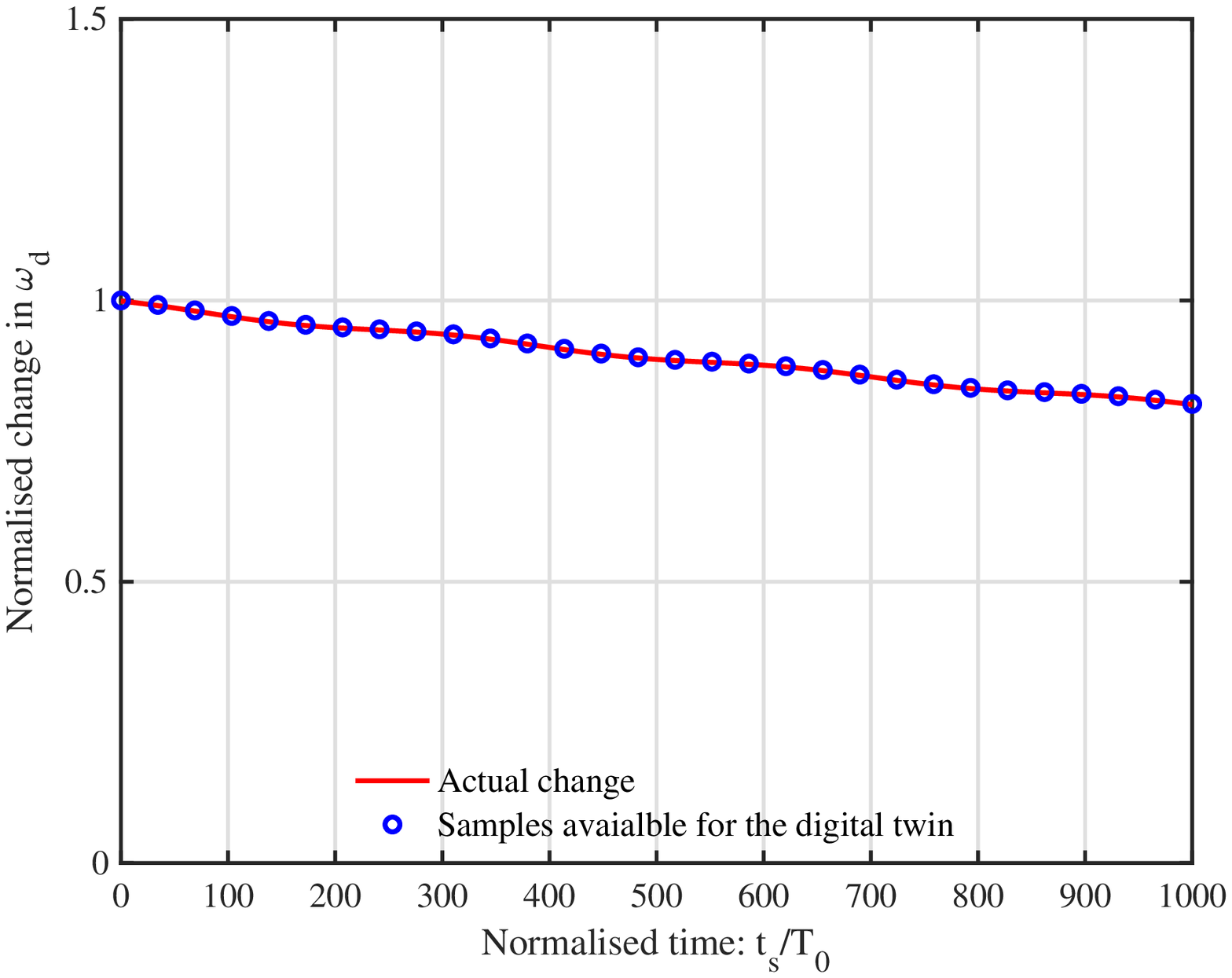}}
  \subfigure[GP assisted digital twin with clean data]{\includegraphics[width = 0.48\textwidth]{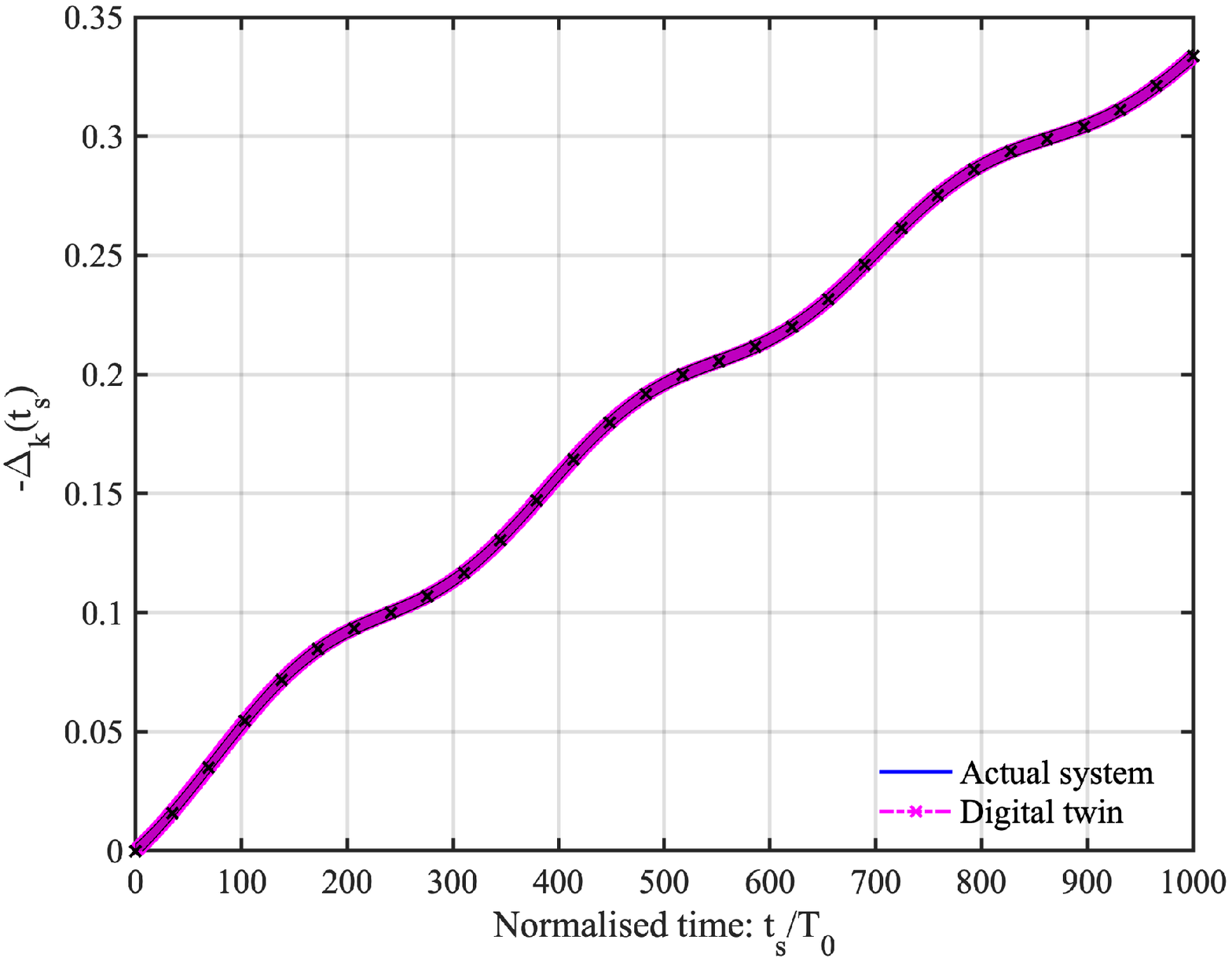}}
  \caption{Changes in the (damped) natural frequency and the GP based digital twin obtained using the exact data plotted as a function of the normalised slow time $t_s / T_0$. Using Bayesian information criteria, `linear' based function and `ARD Rational quadratic' covariance functions are selected. The estimated hyperparameters from Bayesian optimisation are $\bm \beta = [0.1804, 0.1009]$ and $\bm \theta = [0.3729, 3.48\times 10^8, 0.0128]$.}
  \label{fig:gp_dt_stiffness_evol}
\end{figure}

To illustrate the applicability of GP based digital twin with only stiffness evolution, an SDOF system with nominal damping ratio $\zeta_0 = 0.05$ is considered. We assume that the sensor data is transmitted intermittently with a certain regular time interval. For simulating the variation in the natural frequency, we consider the change in the stiffness property of the system shown in \autoref{fig:property_changes}. \autoref{fig:gp_dt_stiffness_evol}(a) shows the actual change in the damped natural frequency of the system with time. The data-points available for the digital twin are also shown. At this stage, it is important to emphasise that the frequency of data availability depends on several factors such as the bandwidth of the wireless transmission system, cost of data collection, {\it etc}. In this case, we have considered 30 data-points. \autoref{fig:gp_dt_stiffness_evol}(b) shows the GP based digital twin model with clean data.We observe that the GP based digital twin perfectly captures the time evolution of $\Delta_k$ with utmost confidence. However, we will like to emphasise that this is an imaginary scenario as it is almost impossible to have measurements with no noise.

\begin{figure}[ht!]
  \centering
  \subfigure[30 observations and $\sigma_{\theta} = 0.005$]{\includegraphics[width=0.48\textwidth]{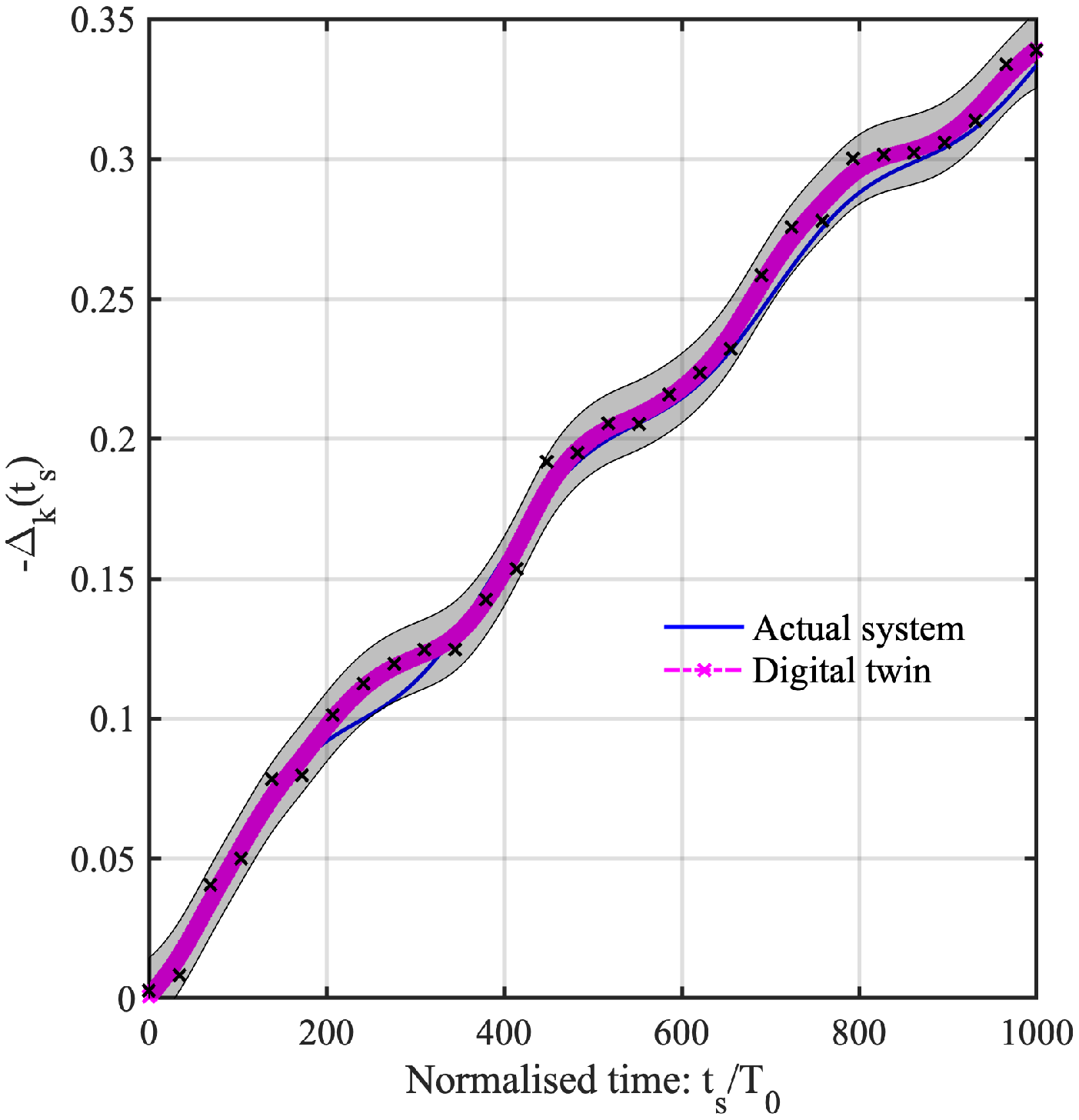}}
  \subfigure[30 observations and $\sigma_{\theta} = 0.015$]{\includegraphics[width=0.48\textwidth]{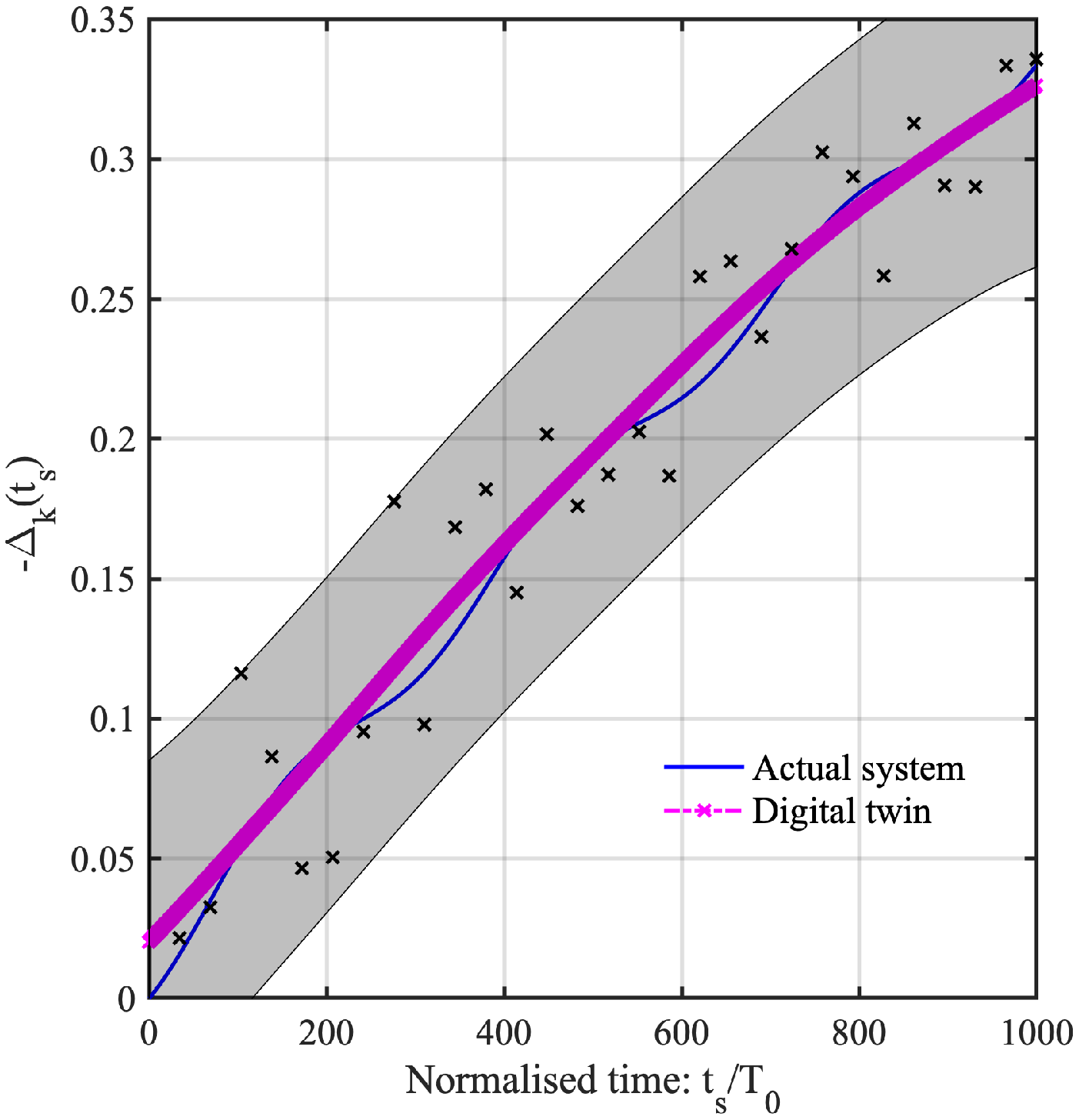}}
  \subfigure[30 observations and $\sigma_{\theta} = 0.025$]{\includegraphics[width=0.48\textwidth]{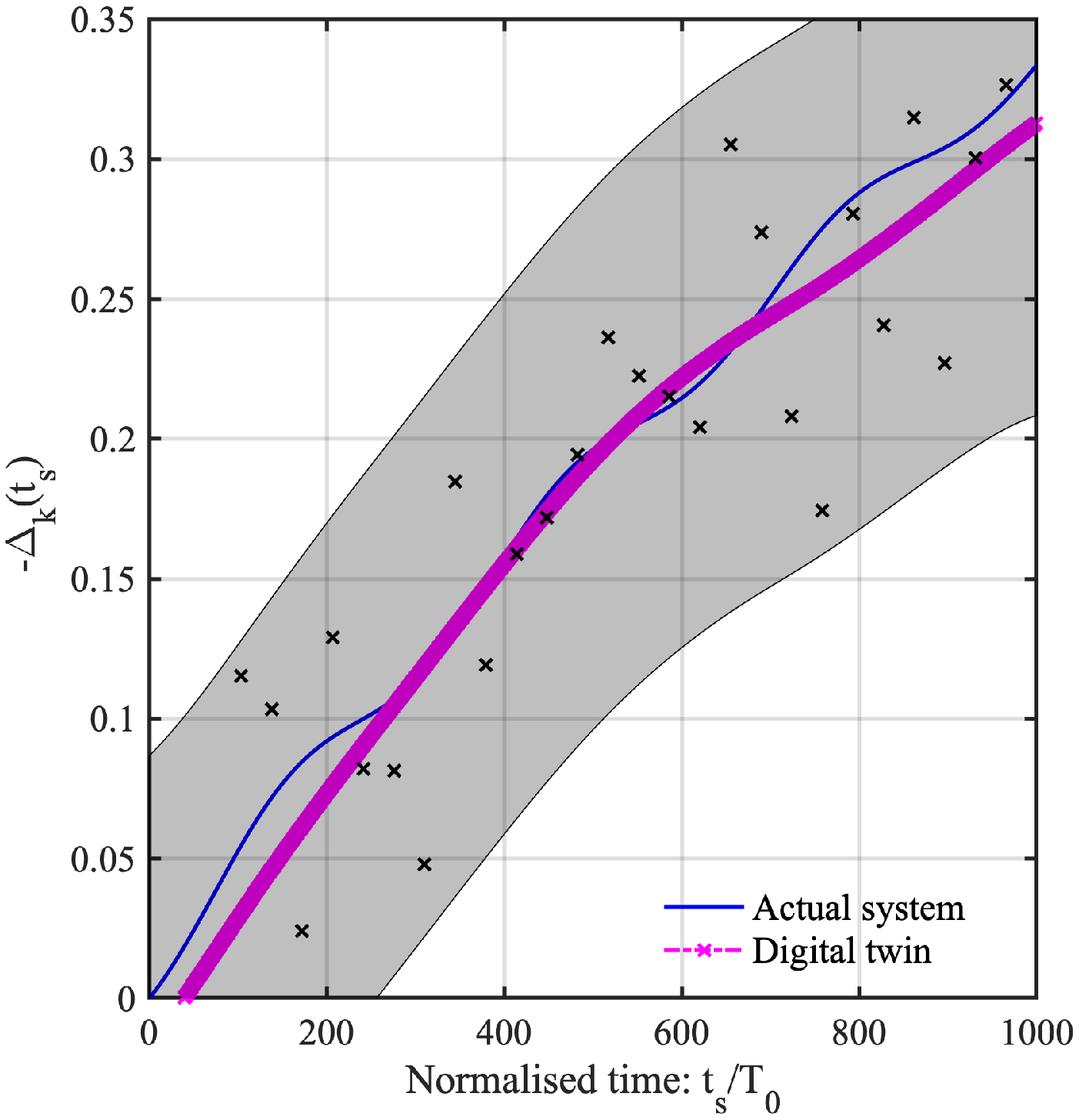}}
  \subfigure[100 observations and $\sigma_{\theta} = 0.025$]{\includegraphics[width=0.48\textwidth]{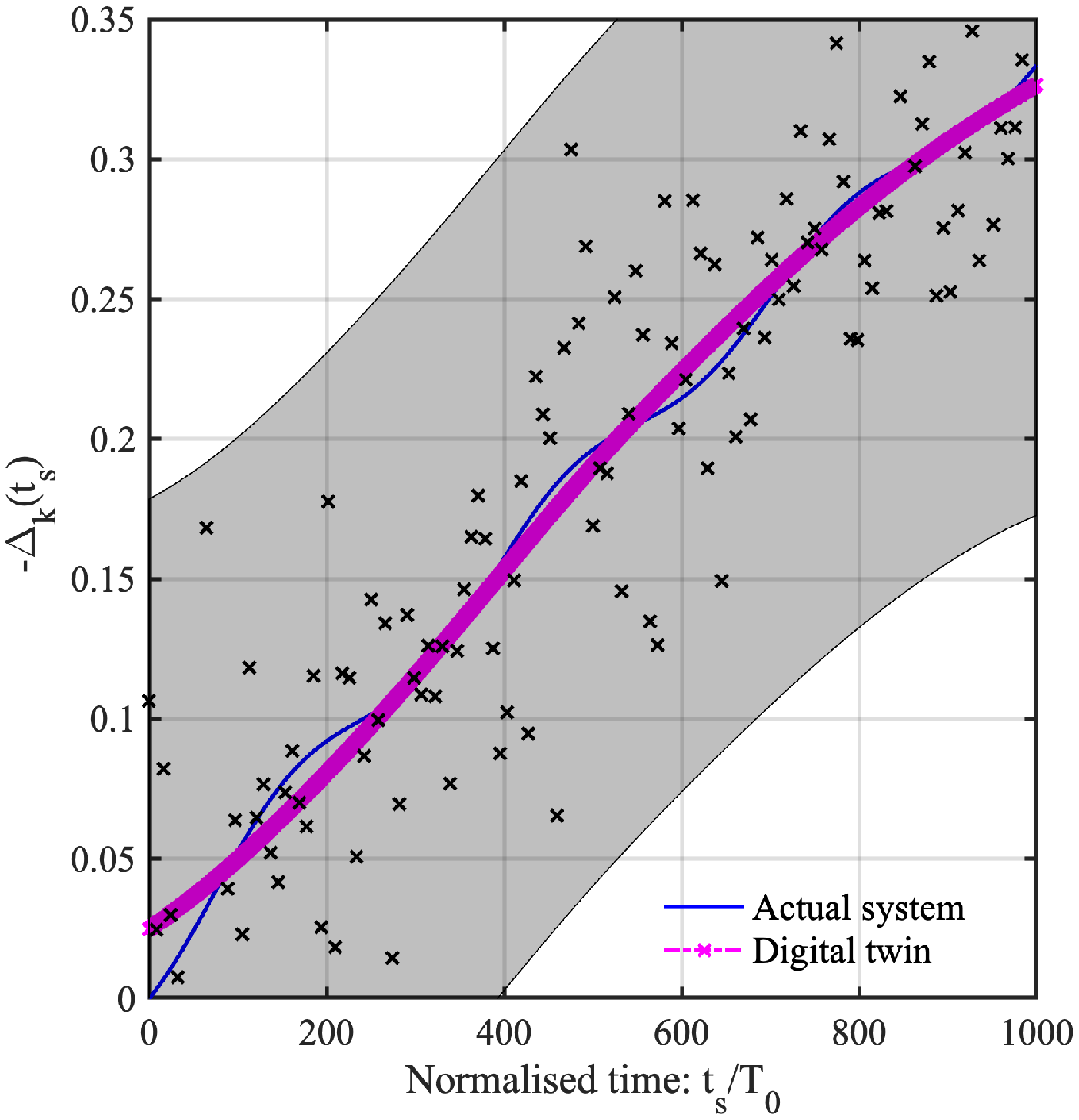}}
  \caption{GP based digital twin constructed with noisy data as a function of the normalised slow time $t_s / T_0$. For (a) - (c), Bayesian information criteria selects `constant' basis function. `ARD Rational quadratic', `ARD squared exponential' and `ARD Matern 3/2' covariance kernels are selected for cases (a) to (c) respectively. For (d), Bayesian information criteria yields `Constant' basis with $\beta = 0.1768$ and `ARD Matern 3/2' covariance kernel with $\bm \theta = [2.7298 \times 10^3, 0.3402]$. The shaded plot depicts the 95\% confidence interval.}
  \label{fig:gp_dt_stiffness_evol_noisy}
\end{figure}

Now we consider a more realistic case where the data collected is corrupted by noise. For illustration purposes, a zero-mean Gaussian white noise with three different standard deviations is considered: (a) $\sigma_{\theta} = 0.005$, (b) $\sigma_{\theta} = 0.015$ and (c) $\sigma_{\theta} = 0.025$. Figs. \ref{fig:gp_dt_stiffness_evol_noisy}(a)--(c) show the GP based digital twin model corresponding to the three cases. For $\sigma_{\theta} = 0.005$, the GP based digital twin successfully captures the evolution of stiffness with time. With the increase in the noise level, a slight deterioration in the performance of the GP based digital twin is observed. Nonetheless, even when $\sigma_{\theta} = 0.025$, the GP based digital twin captures the time-evolution with high accuracy. An additional advantage of GP based digital twin resides in its capability to capture the uncertainty arising due to limited data and noise. This is illustrated by a shaded plot in \autoref{fig:gp_dt_stiffness_evol_noisy}. We observe that with increase in noise level, the uncertainty also increases.

To illustrate the importance of having more data, a case with 100 data-points is shown in \autoref{fig:gp_dt_stiffness_evol_noisy}(d). We observe that with the increase in the number of observations, the GP based digital twin can capture the evolution of stiffness in a more accurate manner even for the highest level of noise in the data.

\subsection{Digital twin via mass evolution}\label{subsec:dt_mass}

\subsubsection{Formulation}
In the second case, we consider that the time-evolution of the natural frequency is due to the change in the mass of the system.
Subsequently, the equation of motion of the digital twin can be represented as
\begin{equation}\label{eq:dt_mass}
    m_s\left(t_s\right)\frac{\text d^2u\left(t\right)}{\text dt^2} + c_0 \frac{\text du\left(t\right)}{\text dt} + k_0u\left(t\right) = f\left(t\right),
\end{equation}
where
\begin{equation}\label{eq:mass_evol_dt}
    m_s \left(t_s\right) = m_0 \left( 1 + \Delta _m \left(t_s\right)\right).
\end{equation}
Similar to case 1, we assume
\begin{equation}\label{eq:gp_mass_evol}
    \Delta _m \left(t_s\right) \approx \Delta _{\hat m} \left(t_s\right) \sim \mathcal{GP}\left(\mu_m \left(t_s\right), \kappa_{\hat m} \left(t_s\right) \right),
\end{equation}
substitute Eqs. (\ref{eq:mass_evol_dt}) and (\ref{eq:gp_mass_evol}) into
\autoref{eq:dt_mass} and solve it to obtain the natural frequency of
the system
\begin{equation}\label{eq:nat_freq_mass}
    \lambda_{s_{1,2}} \left(t_s\right) = - \omega_s\left(t_s\right)\zeta_s\left(t_s\right) \pm \text i \omega_{d_s}\left(t_s\right),
\end{equation}
where
\begin{subequations}
    \begin{equation}
        \omega_s\left(t_s\right) = \frac{\omega_0}{\sqrt{1 + \Delta_{\hat m}\left(t_s\right)}},
    \end{equation}
    \begin{equation}
        \zeta_s\left(t_s\right) = \frac{\zeta_0}{\sqrt{1 + \Delta_{\hat m}\left(t_s\right)}}\;\; \text{and}
    \end{equation}
    \begin{equation}
        \omega_{d_s}\left(t_s\right) = \omega_s \left(t_s\right)\sqrt{1 - \zeta_s^2\left(t_s\right)}.
    \end{equation}
\end{subequations}
are the evolution of natural frequency, damping ratio and damped natural frequency of the digital twin.
Again, following a procedure similar to case 1, we obtain
\begin{equation}\label{eq:mass_cf}
\begin{split}
    \Delta_{\hat m}\left(t_s\right) = & \frac{-2\tilde d_2\left(t_s\right)^2 + 4\tilde d_2\left(t_s\right)\sqrt{1 - \zeta_0^2} - 1 + 2\zeta_0^2 }{2 \left( - \tilde d_2 \left(t_s\right) + \sqrt{1 - \zeta_0^2} \right)^2} \\
   & + \frac{\sqrt{1 - 4\tilde d_2\left(t_s\right)^2 \zeta_0^2 + 8 \tilde d_2\left(t_s\right) \sqrt{1 - \zeta_0^2}\zeta_0^2 - 4\zeta_0^2 + 4 \zeta_0^4}}{2 \left( - \tilde d_2 \left(t_s\right) + \sqrt{1 - \zeta_0^2} \right)^2}
\end{split}
\end{equation}
where $\tilde d_2 \left(t_s\right)$ is the equivalent of $\tilde d_1$ for the mass evolution case. Similar to case 1, $\Delta_{\hat m}\left(t_s\right)$ obtained using \autoref{eq:mass_cf} will be noisy in nature. We consider this noisy estimates of $\Delta_{\hat m}\left(t_s\right)$ to be the training outputs and $t_s$ to be the training inputs for estimating the hyperparameters, $\bm \theta$, of the GP defined in \autoref{eq:gp_mass_evol}. This is achieved by maximising the likelihood of the data.
The parameter settings for solving the optimization problem are kept similar as before.
For determining the best mean function and covariance function, Bayesian information criteria as before has been adopted. Once the hyperparameter, $\bm \theta$, has been estimated, the digital twin model defined in \autoref{eq:dt_mass} is completely deciphered.

\subsubsection{Numerical illustration}
\begin{figure}[htbp!]
    \centering
    \subfigure[Changes in the (damped) natural frequency]{\includegraphics[width = \textwidth]{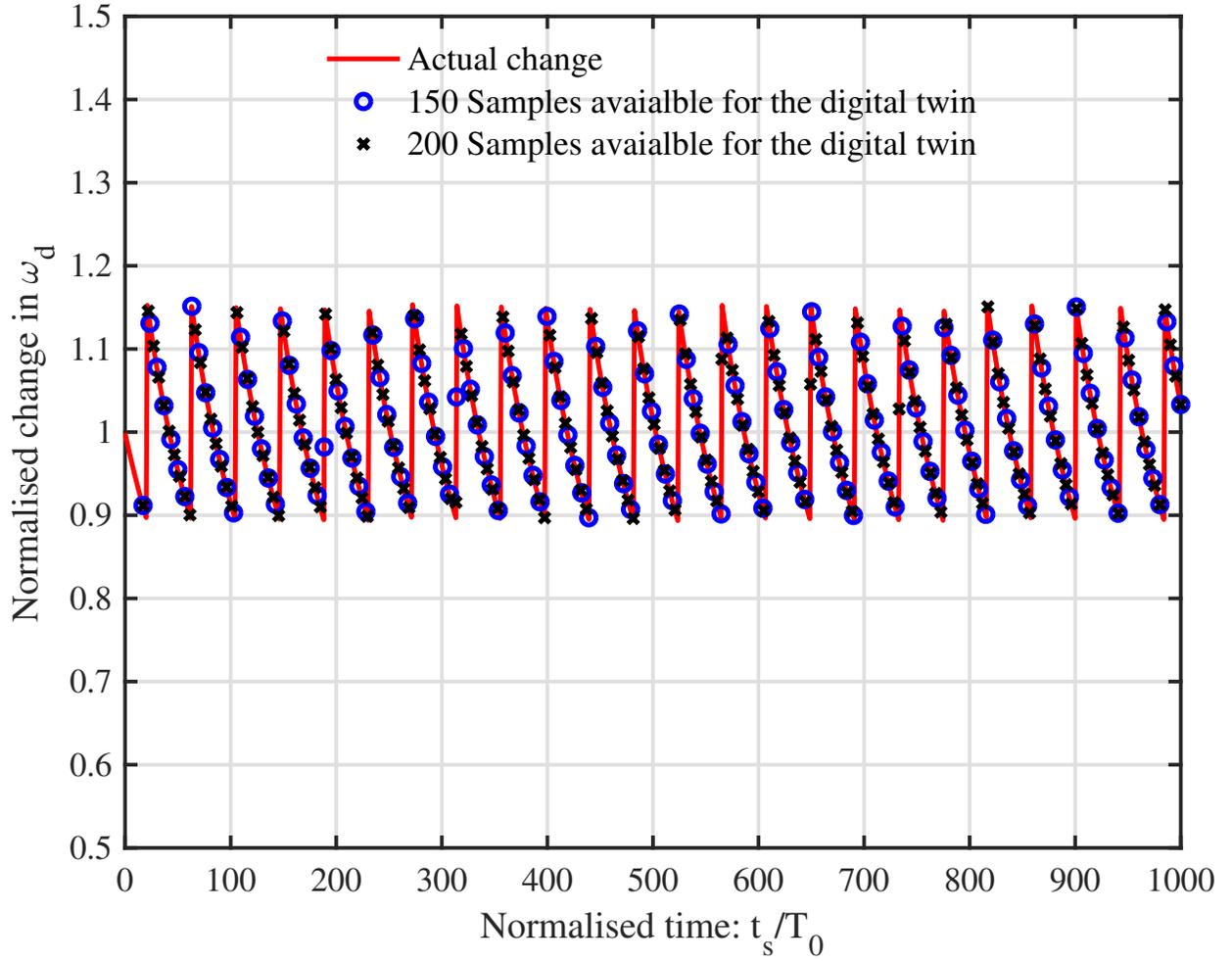}}
    \subfigure[GP based digital twin]{\includegraphics[width = \textwidth]{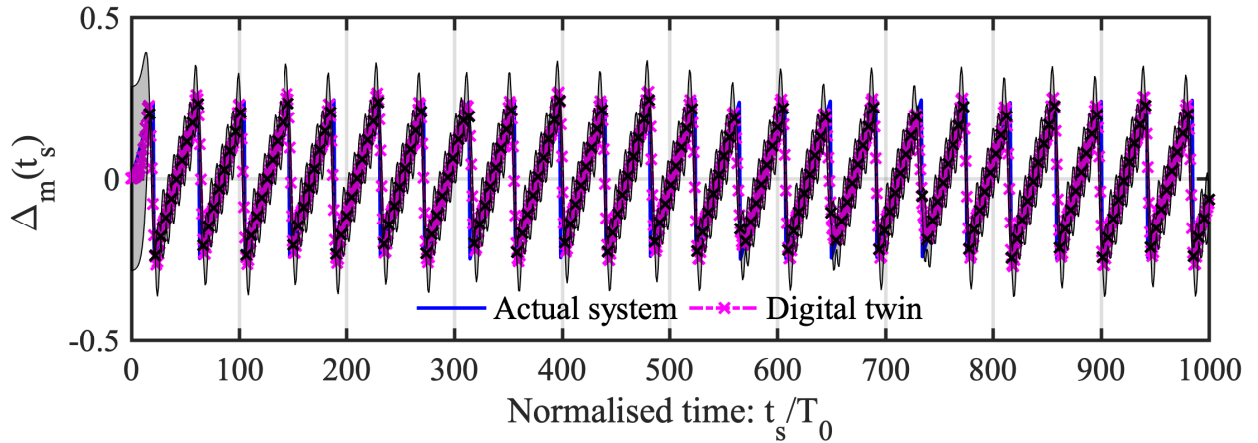}}
    \caption{Changes in the (damped) natural frequency and the GP based digital twin obtained using the exact data plotted as a function of the normalised slow time $t_s / T_0$. Using Bayesian information criteria, `Constant' basis function and `ARD Matern 5/2' covariance functions are selected. The estimated hyperparameters from Bayesian optimisation are $ \beta = 0.0$ and $\bm \theta = [4.1532, 0.1429]$. The shaded plot depicts the 95\% confidence interval.}
    \label{fig:gp_dt_mass_evol}
\end{figure}

To illustrate the applicability of the GP based digital twin model for mass evolution, we revisit the SDOF example considered for stiffness evolution. For simulating the variation in the natural frequency, we consider the change in the mass of the system shown in \autoref{fig:property_changes}.  \autoref{fig:gp_dt_mass_evol}(a) shows the actual change in the damped natural frequency of the system with time. The data-points available for the digital twin are also shown. Note that the evolution of mass with time is more complex and hence, unlike the stiffness evolution case, only 30 observations is inadequate. Therefore, we have assumed access to more data points in this case. \autoref{fig:gp_dt_mass_evol}(b) shows the GP based digital twin constructed from clean data. It is observed that the GP based digital twin is able to capture the time evolution of mass with high accuracy. Also the uncertainty due to limited data is adequately captured. However, as already stated, this is an unrealistic case as for a practical scenario, data collected will always be corrupted by some form of noise.

\begin{figure}
    \centering
        \includegraphics[width = \textwidth]{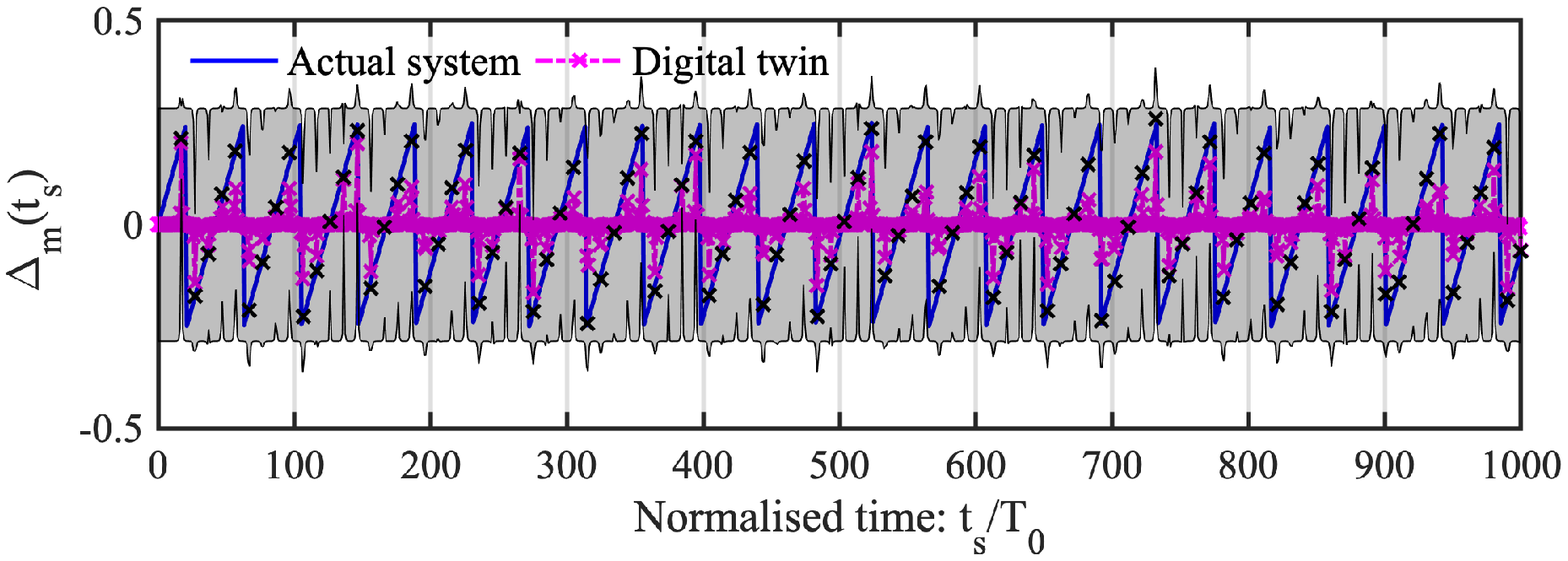}
    \caption{GP based digital twin obtained using 100 noisy data with $\sigma_{\theta} = 0.005$, plotted as a function of the normalised slow time $t_s / T_0$. The shaded plot depicts the 95\% confidence interval.}
    \label{fig:gp_dt_mass_evol_noisy1}
\end{figure}

\begin{figure}[ht!]
    \centering
    \subfigure[$\sigma_{\theta} = 0.005$]{
    \includegraphics[width = \textwidth]{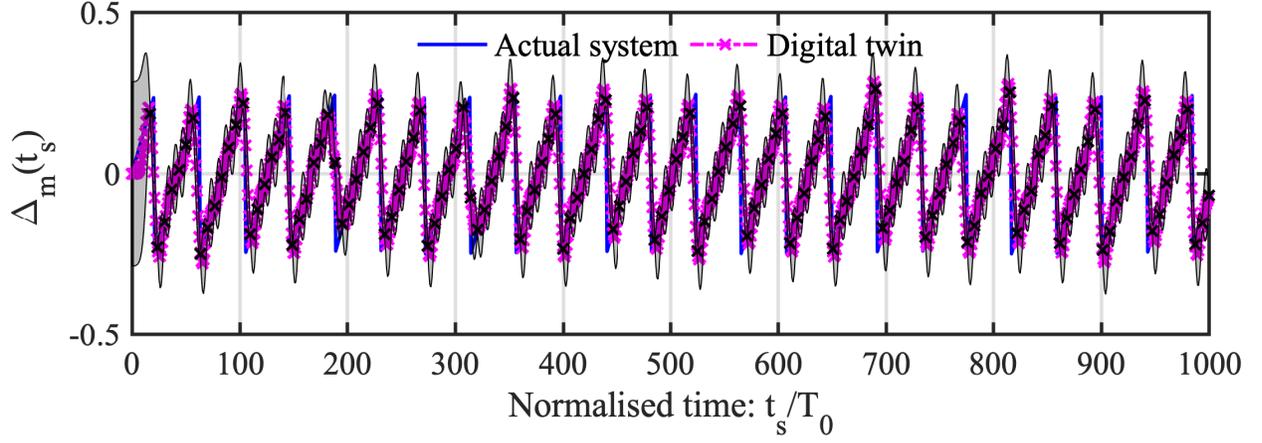}}
    \subfigure[$\sigma_{\theta} = 0.015$]{
    \includegraphics[width = \textwidth]{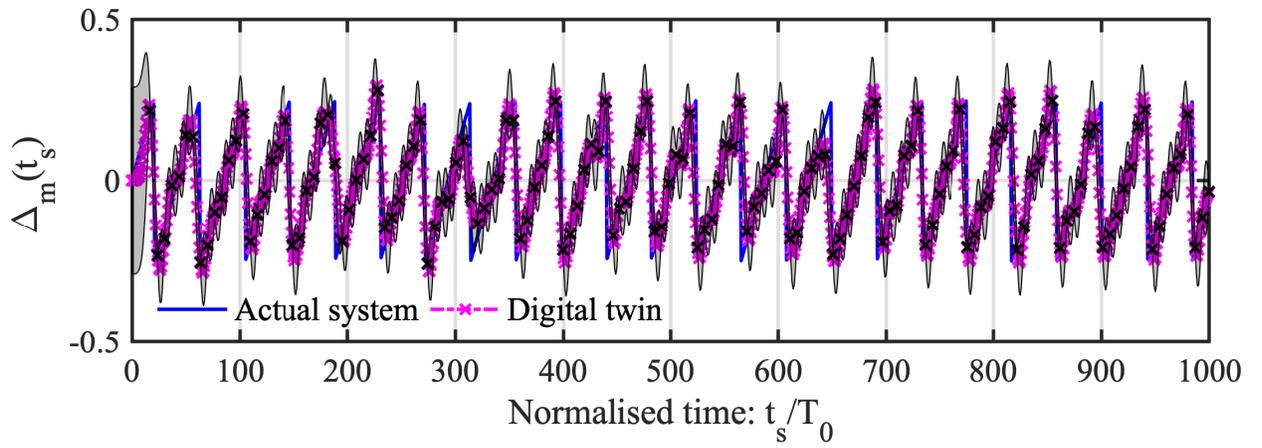}}
    \subfigure[$\sigma_{\theta} = 0.025$]{
    \includegraphics[width = \textwidth]{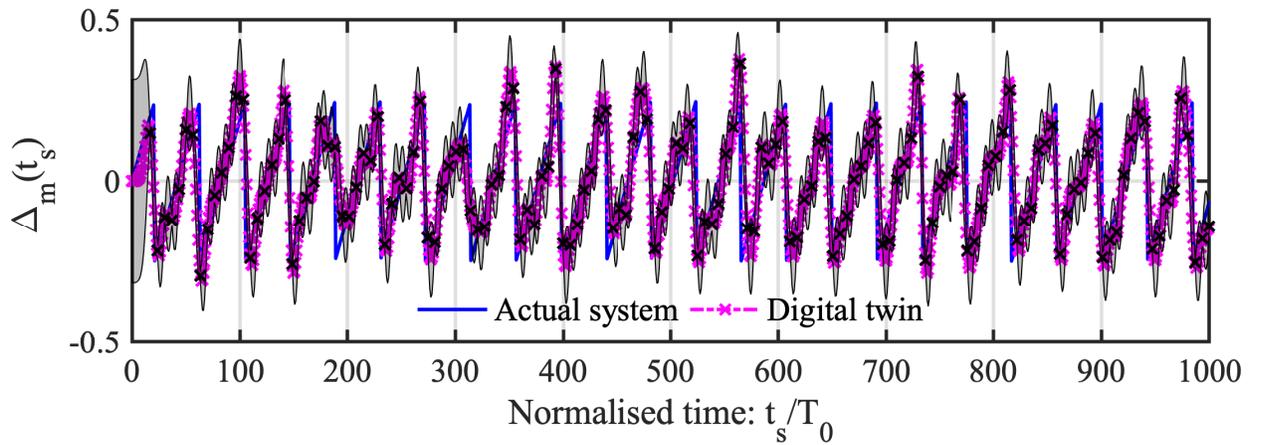}}
    \caption{GP based digital twin obtained using 150 noisy data plotted as a function of the normalised slow time $t_s / T_0$. For (a) - (c), Bayesian information criteria selects `Constant' basis and `ARD Matern 5/2' covariance kernel. The shaded plot depicts the 95\% confidence interval.}
    \label{fig:gp_dt_mass_evol_noisy2}
\end{figure}

\begin{figure}ht!]
    \centering
    \includegraphics[width = \textwidth]{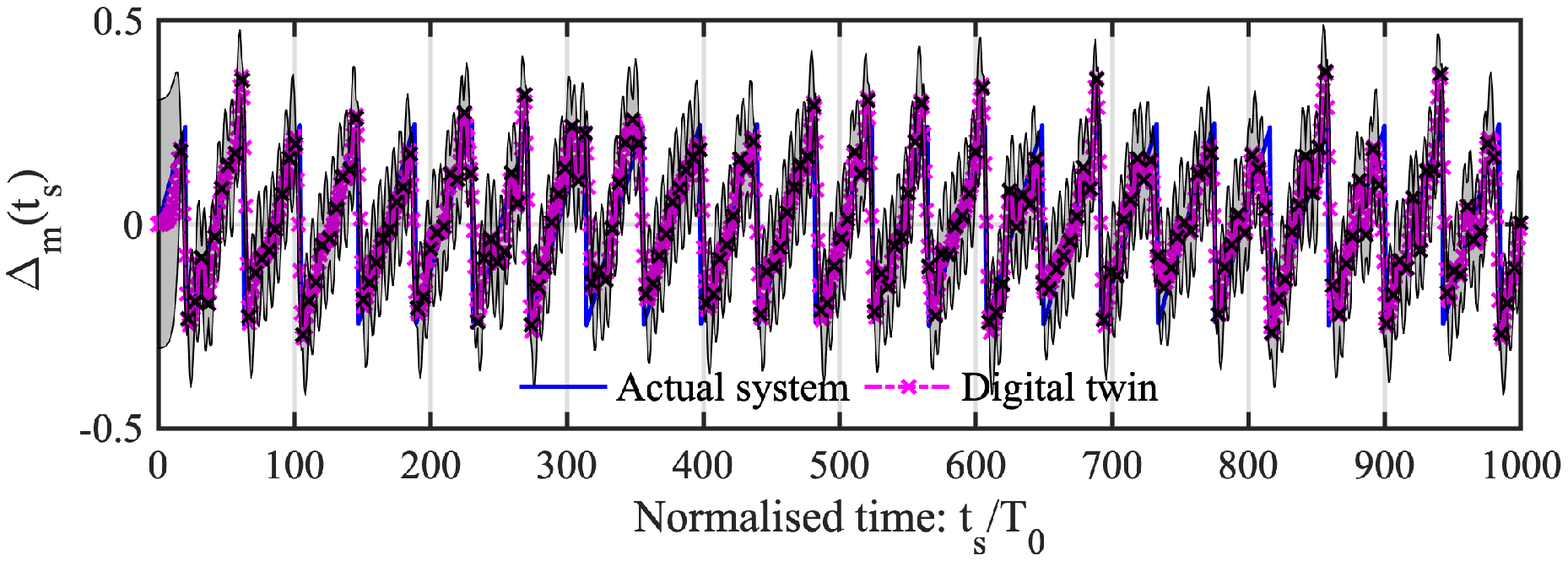}
    \caption{GP based digital twin obtained using 200 noisy data with $\sigma_{\theta} = 0.025$ plotted as a function of the normalised slow time $t_s / T_0$. Bayesian information criteria yields 'Constant' basis with $\beta = 0.0$ and `ARD Matern 5/2' covariance kernel with $\bm \theta = [4.1242, 0.1527]$.}
    \label{fig:gp_dt_mass_evol_noisy3}
\end{figure}

Next, we consider a more realistic case where the data collected is corrupted by noise. \autoref{fig:gp_dt_mass_evol_noisy1} shows the GP based digital twin model trained with 100 noisy observations and $\sigma_{\theta} = 0.005$. Some discrepancy between the GP based digital twin and the physical twin is observed. For improved performance, we increase the number of observations to 150. The GP based digital twin corresponding to the three noise level are shown in \autoref{fig:gp_dt_mass_evol_noisy2}(a) -- (c). In this case, the GP based digital twin yields accurate result. Lastly, \autoref{fig:gp_dt_mass_evol_noisy3} shows  GP based digital twin with 200 samples and $\sigma_{\theta} = 0.025$. This yields the best result with accurate time evolution of mass and uncertainty quantification. Overall, this example illustrates the importance of sampling rate and denoising of data for digital twin technology.

\subsection{Digital twin via mass and stiffness evolution}\label{subsec:dt_mass_stiffness}

\subsubsection{Formulation}
As the last case, we consider simultaneous evolution of the mass and stiffness.
The governing differential equation for this case is represented as
\begin{equation}\label{eq:mass_stiff_dt}
    m_s\left(t_s\right)\frac{\text d^2u\left(t\right)}{\text dt^2} + c_0\frac{\text du\left(t\right)}{\text dt} + k_s\left(t_s\right)u\left(t\right) = f\left(t\right),
\end{equation}
where $k_s\left(t_s\right)$ and $m_s\left(t_s\right)$ are represented by Eqs. (\ref{eq:stiff_evol}) and (\ref{eq:mass_evol_dt}), respectively.
We represent $\Delta_{m}\left(t_s\right)$ and $\Delta_{ k}\left(t_s\right)$
by a multi-output GP \cite{Bilionis2013multi,alvarez2009sparse},
\begin{equation}\label{eq:mogp}
    \left[ \Delta_{m}\left(t_s\right), \Delta_{k}\left(t_s\right) \right] \approx \left[ \Delta_{\hat m}\left(t_s\right), \Delta_{\hat k}\left(t_s\right) \right] \sim \mathcal{GP}\left(\bm \mu_{t_s}, \bm \kappa \left(t_s, t_s'; \bm \theta\right)\right),
\end{equation}
Substituting Eqs. (\ref{eq:mogp}), (\ref{eq:stiff_evol}) 
and (\ref{eq:mass_evol_dt}) in \autoref{eq:mass_stiff_dt} and solving, we obtain
\begin{equation}\label{eq:freq}
    \lambda_{s_{1,2}} = -\omega_s \left(t_s\right) \zeta_s\left(t_s\right) \pm \text{i} \omega_{d_s}\left(t_s\right),
\end{equation}
where
\begin{subequations}
    \begin{equation}
        \omega_s \left(t_s\right) = \omega_0 \frac{\sqrt{1 + \Delta_{\hat k}\left(t_s\right)}}{\sqrt{1 + \Delta_{\hat m}\left(t_s\right)}}
    \end{equation}
    \begin{equation}
        \zeta_s\left(t_s\right) = \frac{\zeta_0}{\sqrt{1 + \Delta_{\hat m}\left(t_s\right)}\sqrt{1 + \Delta_{\hat k}\left(t_s\right)}}\;\;\text{and}
    \end{equation}
    \begin{equation}
        \omega_{d_s} \left(t_s\right) = \omega_{s} \left(t_s\right)\sqrt{1 - \zeta_s^2}.
    \end{equation}
\end{subequations}
Note that unlike the previous two cases, it is not possible to compute the hyperparameters of the GPs based on measurements of damped natural frequencies only.
Therefore, we take a different route by considering the real and imaginary
part in \autoref{eq:freq} separately.
Following Ganguli and Adhikari \cite{Ganguli2020}, it can be shown that
\begin{subequations}\label{eq:delta_mk}
    \begin{equation}
        \Delta_{\hat m} \left(t_s\right) = - \frac{\tilde d_{\mathcal R} \left(t_s\right)}{\zeta_0 + \tilde d_{\mathcal R} \left(t_s\right)},
    \end{equation}
    \begin{equation}
        \Delta_{\hat k} \left(t_s\right) = \frac{\zeta_o \tilde d_{\mathcal R}^2 \left(t_s\right) - \left( 1 - 2 \zeta_0^2 \right) \tilde d_{\mathcal I} \left(t_s\right) + \zeta_0^2 \tilde d_{\mathcal I}^2 \left(t_s\right)}{\zeta_0 + \tilde d_{\mathcal R}\left(t_s\right)},
    \end{equation}
\end{subequations}
where $\tilde d_{\mathcal R}\left(t_s\right)$ and $\tilde d_{\mathcal I}\left(t_s\right)$, as before, are distance measures
\begin{equation}
    \tilde d_{\mathcal R}\left(t_s\right) = \frac{d_{\mathcal R}\left(t_s\right)}{1 + \Delta_{\hat m}\left(t_s\right)},\;\;\;\;\; \tilde d_{\mathcal I}\left(t_s\right) = \sqrt{1 - \zeta_0^2} - \frac{\sqrt{\left( 1 + \Delta _{\hat k}\left(t_s\right) \right)\left( 1 + \Delta _{\hat m}\left(t_s\right) \right) - \zeta_0^2}}{1 + \Delta_{\hat m}\left(t_s\right)}.
\end{equation}
We emphasise that \autoref{eq:delta_mk} provides estimates $\Delta _{\hat m}\left(t_s\right)$and $\Delta _{\hat k}\left(t_s\right)$ based on noisy measurements, and hence, are also noisy. Using this noisy estimates as training outputs and $t_s$ as the training inputs, we estimate the hyperparameters, $\bm \theta$ of the multi-output GP defined in \autoref{eq:mogp}. This is achieved by maximising the likelihood of the data as described in \autoref{sec:gp}.
Same parameter settings as before is considered.
For determining the best mean function and covariance function, Bayesian information criteria is adopted as before. The hyperparameters, $\bm \theta$ once estimated completely defines the digital twin described in \autoref{eq:mass_stiff_dt}.

\subsubsection{Numerical illustration}

\begin{figure}[ht!]
    \centering
    \subfigure[Changes in real part of natural frequency]{\includegraphics[width=0.48\textwidth]{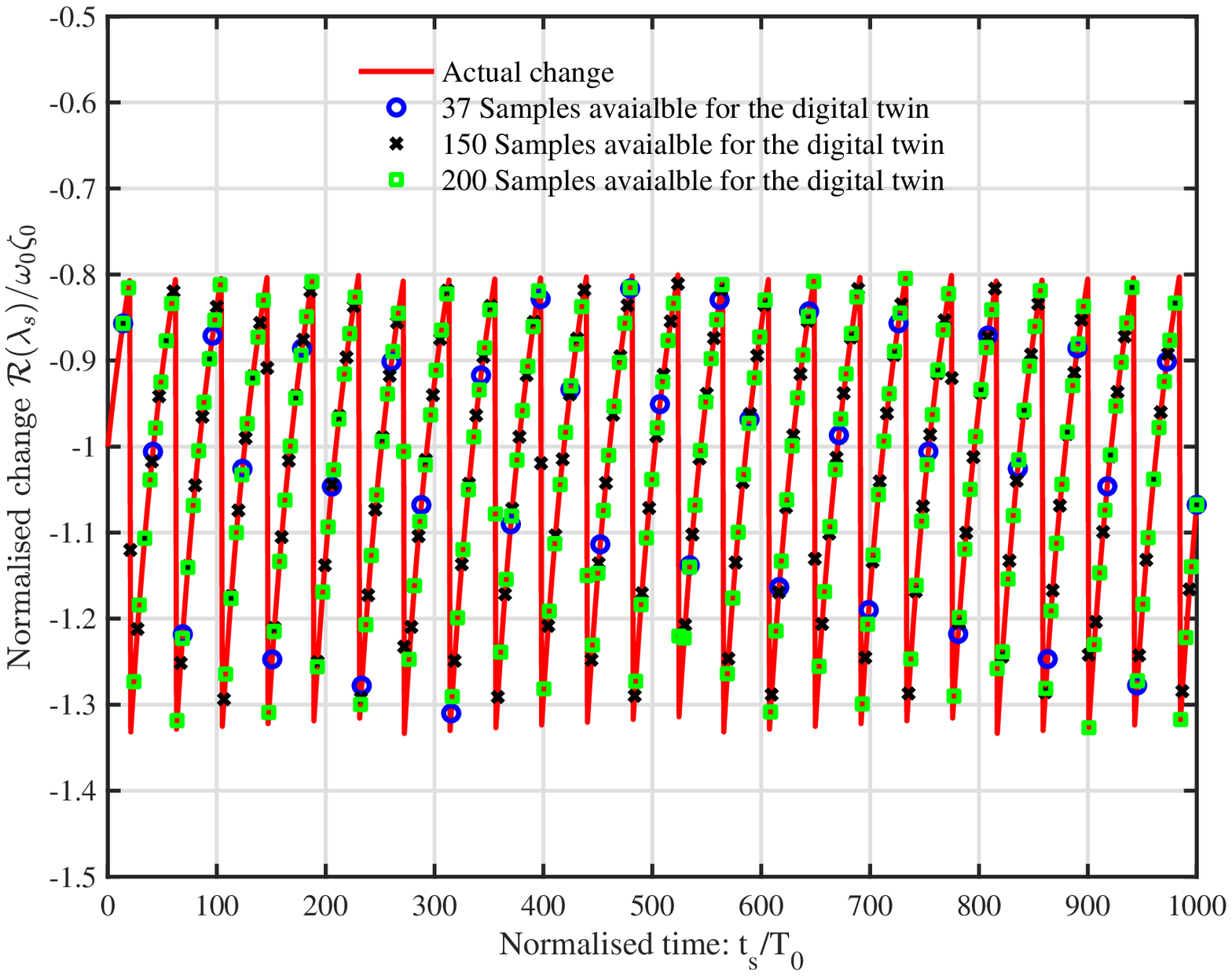}}
    \subfigure[Changes in imaginary part of natural frequency]{\includegraphics[width=0.48\textwidth]{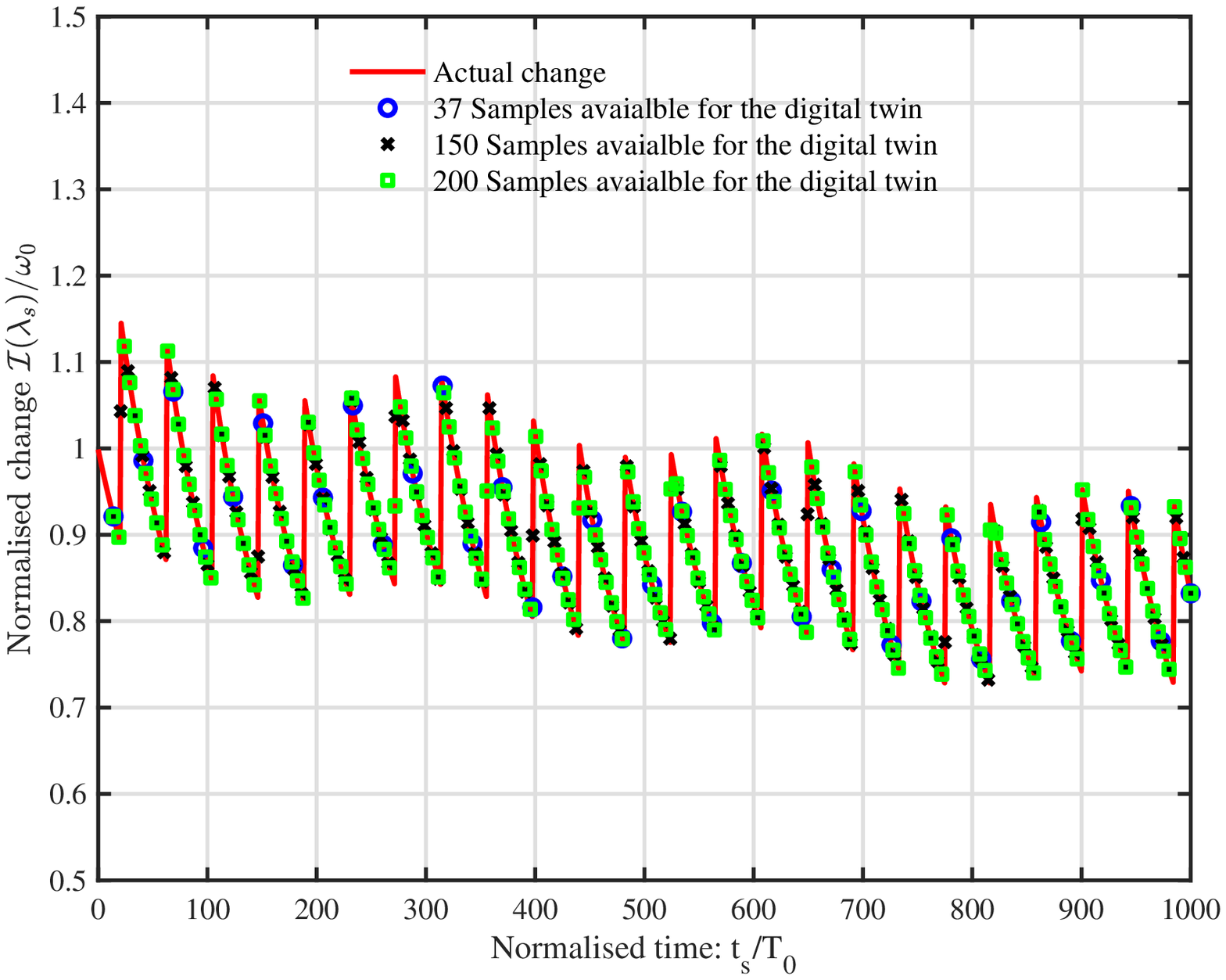}}
    \caption{Normalised changes in the real and imaginary parts of the natural frequency as a function of the normalised slow time $t_s / T_0$. The shaded plot depicts the 95\% confidence interval.}
    \label{fig:data_dt}
\end{figure}

\begin{figure}[ht!]
    \centering
    \subfigure[mass function]{\includegraphics[width=0.8\textwidth]{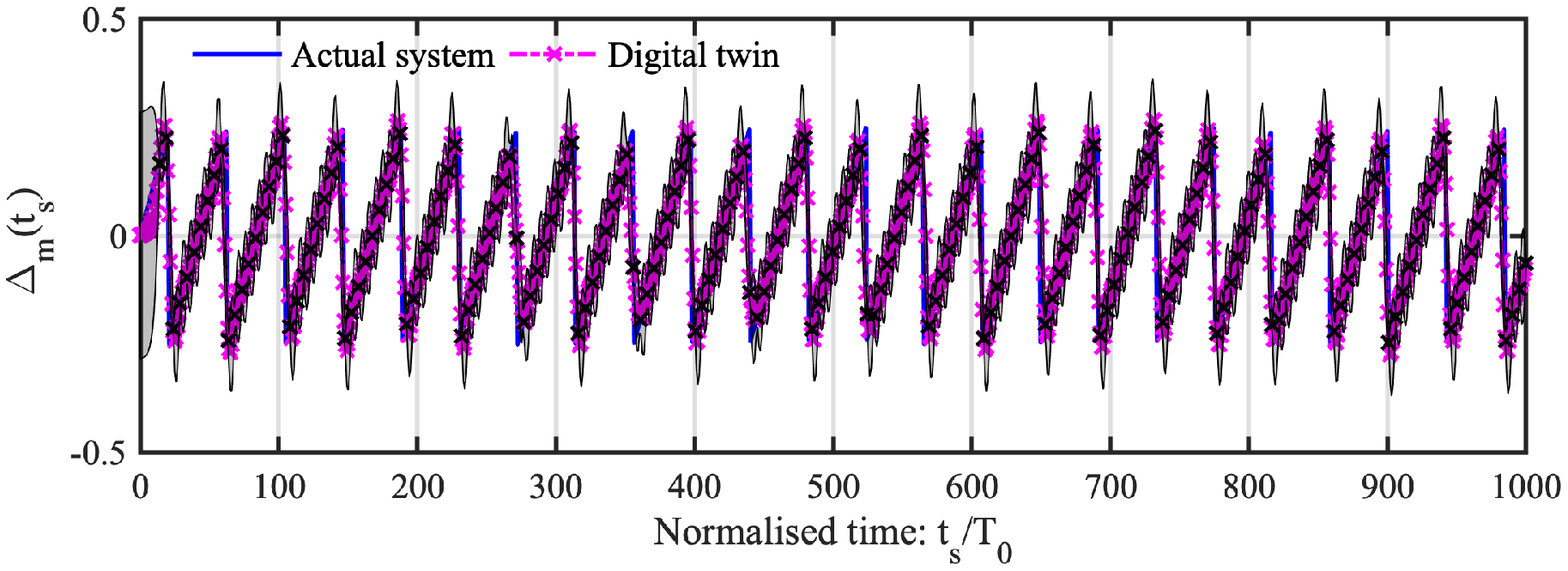}}
    \subfigure[stiffness function]{\includegraphics[width=0.68\textwidth]{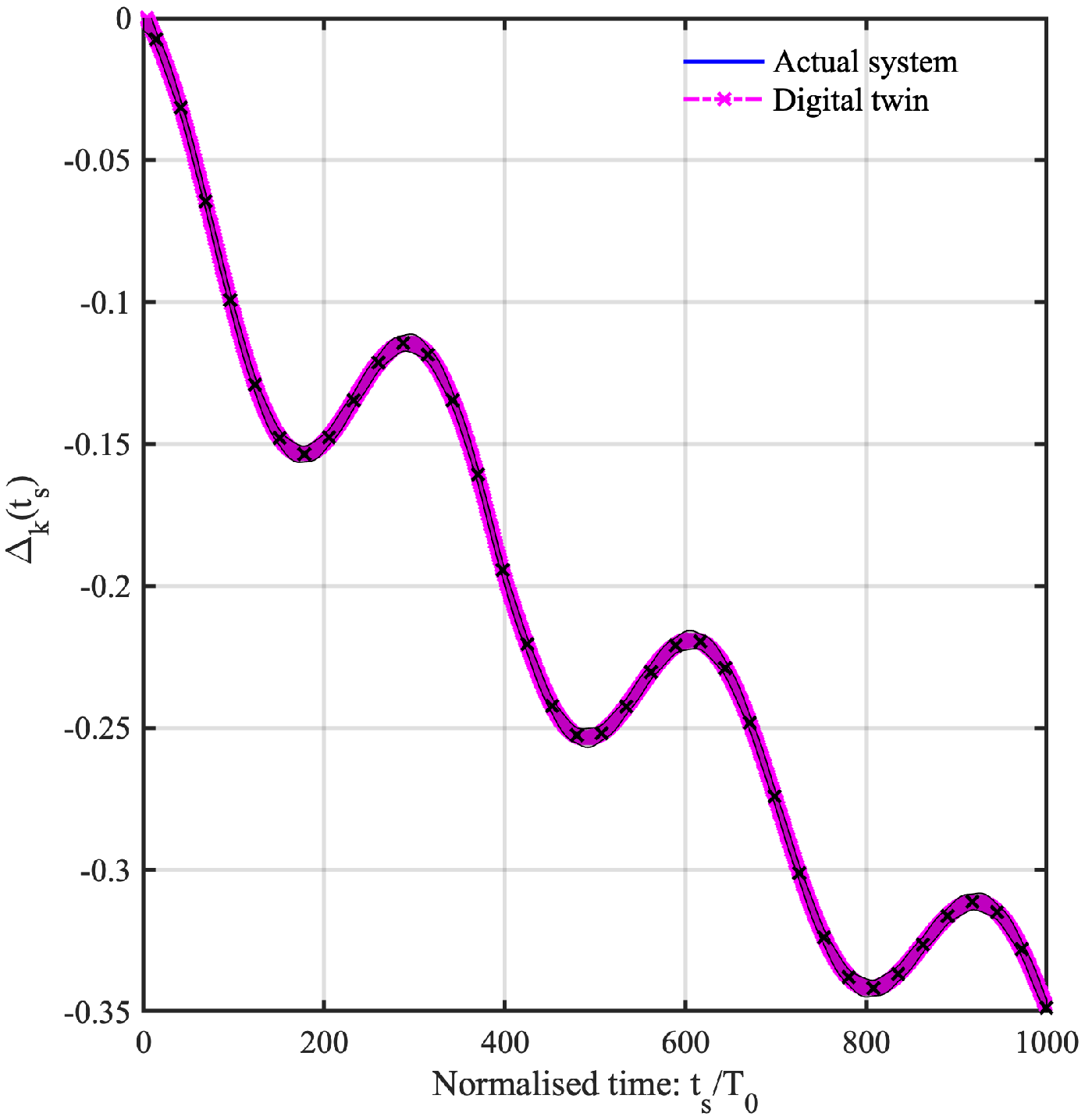}}
    \caption{GP based digital twin obtained from exact data via simultaneous mass and stiffness evolution as a function of the normalised `slow time $t_s / T_0$. Bayesian information criteria yields `Linear' basis and `squared exponential' covariance kernel.}
    \label{fig:dt_mass_stiff_evol}
\end{figure}

We revisit the previously studied SDOF system to illustrate the applicability of GP based digital twin for simultaneous mass and stiffness evolution. For simulating the variation in the natural frequency, we consider the change in the mass of the system shown in \autoref{fig:property_changes}. \autoref{fig:data_dt} shows the actual change in real and imaginary parts of the natural frequency of the system over time. Similar to previous cases, the damping ratio is assumed to be 0.05. \autoref{fig:dt_mass_stiff_evol} shows the GP based digital twin constructed from clean data. The trained GP based digital twin is able to capture the time evolution of mass and stiffness. However, data with no noise is an unrealistic scenario as, even with the most advanced sensors, the data collected will always be noisy \cite{zhang2017new}.

\begin{figure}[ht!]
    \centering
    \subfigure[mass function]{\includegraphics[width=0.8\textwidth]{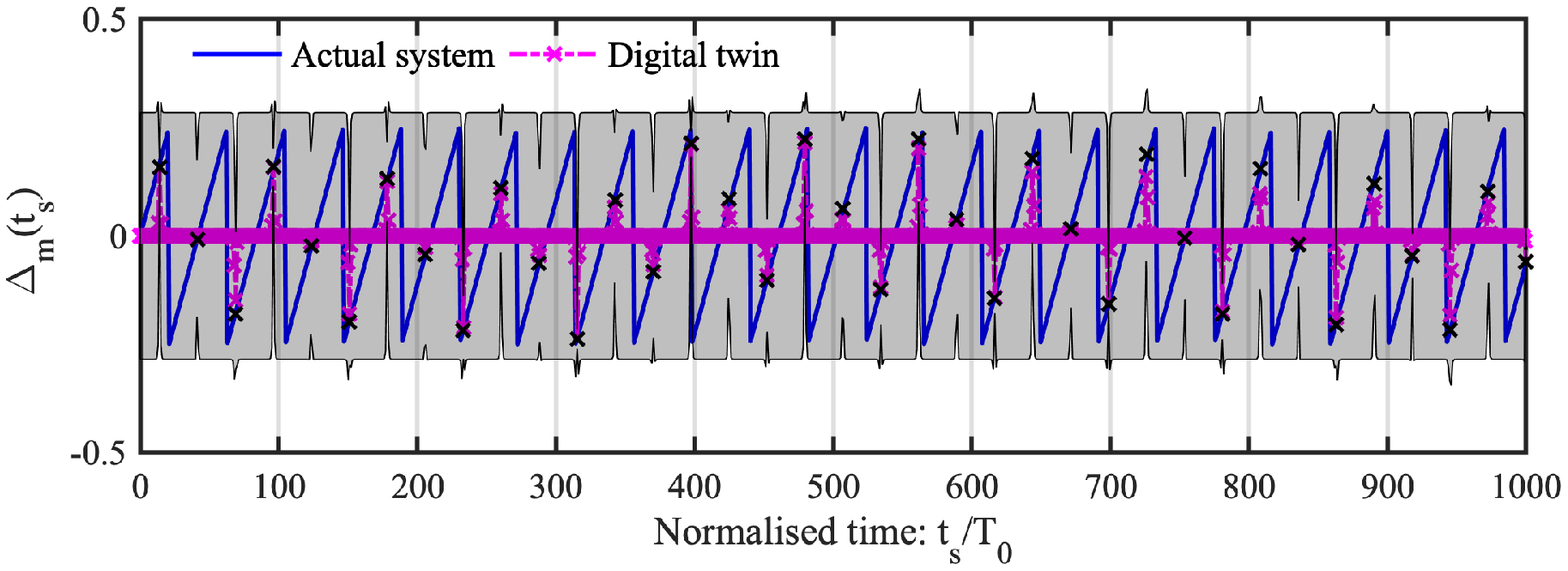}}
    \subfigure[stiffness function]{\includegraphics[width=0.68\textwidth]{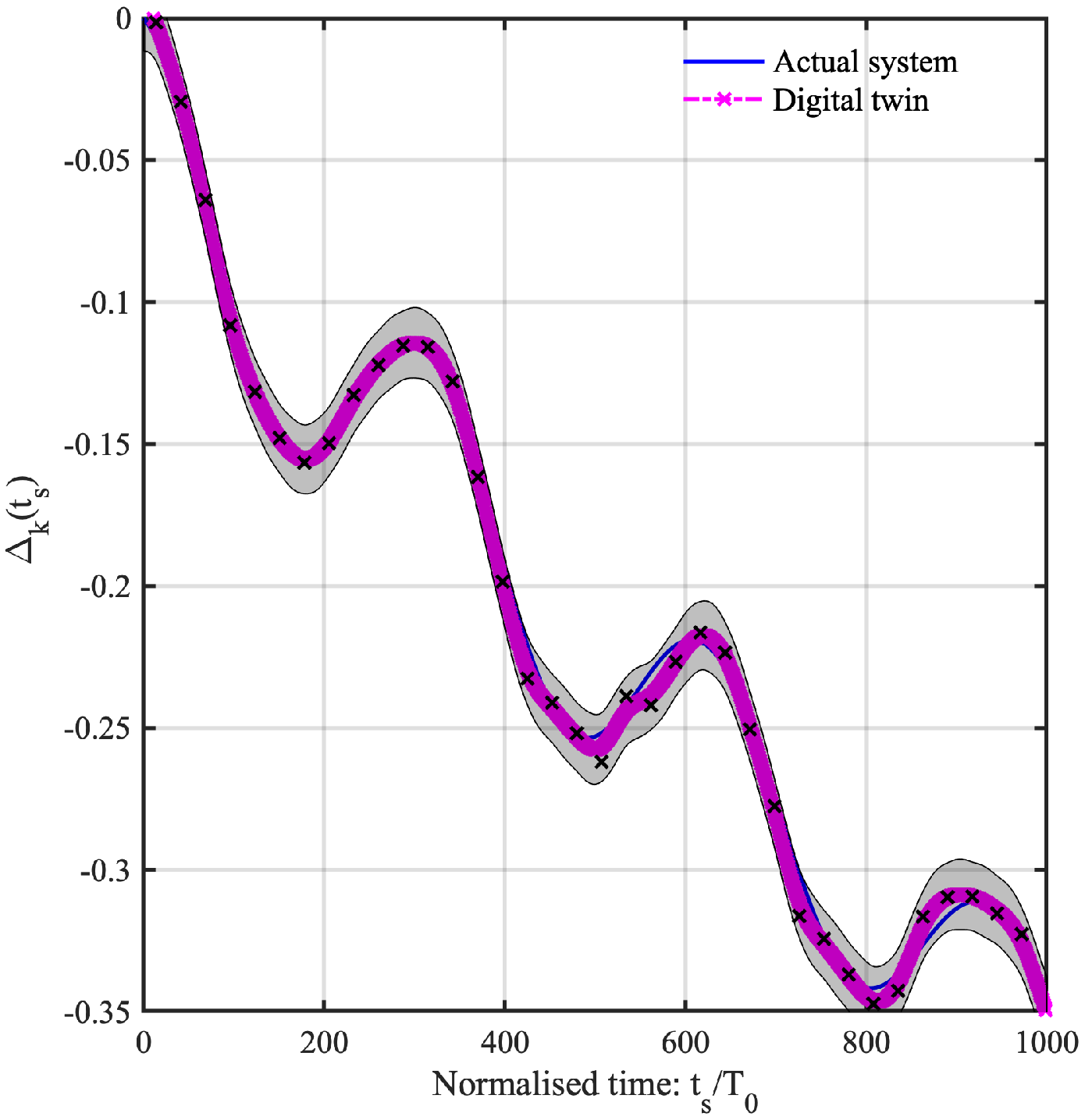}}
    \caption{GP based digital twin obtained from 37 noisy data with $\sigma_{\theta} = 0.005$ via simultaneous mass and stiffness evolution as a function of the normalised slow time $t_s / T_0$. Bayesian information criteria yields `Linear' basis and `squared exponential' covariance kernel. The shaded plot depicts the 95\% confidence interval.}
    \label{fig:gp_dt_mass_stiff_evol_noisy1}
\end{figure}

Next, we consider a more realistic case where the sensor data is corrupted by noise. Similar to the previous two cases, we consider three noise levels. \autoref{fig:gp_dt_mass_stiff_evol_noisy1} shows the mass and stiffness evolution of the GP based digital twin trained with 37 noisy observations.The observations are corrupted by white Gaussian noise with $\sigma_{\theta} = 0.005$. It is observed that the GP based digital twin is able to capture the time evolution of stiffness with high accuracy. However, it fails to capture the mass evolution in an adequate manner. This is expected as the mass evolution function is quite complex and it is unreasonable to expect that with so little data, GP will be able to track the mass evolution. Note that the uncertainty due to limited and noisy data is perfectly captured and hence, the true solution resides within the shaded portion.

\begin{figure}[ht!]
    \centering
    \subfigure[$\sigma_{\theta} = 0.005$]{\includegraphics[width=0.8\textwidth]{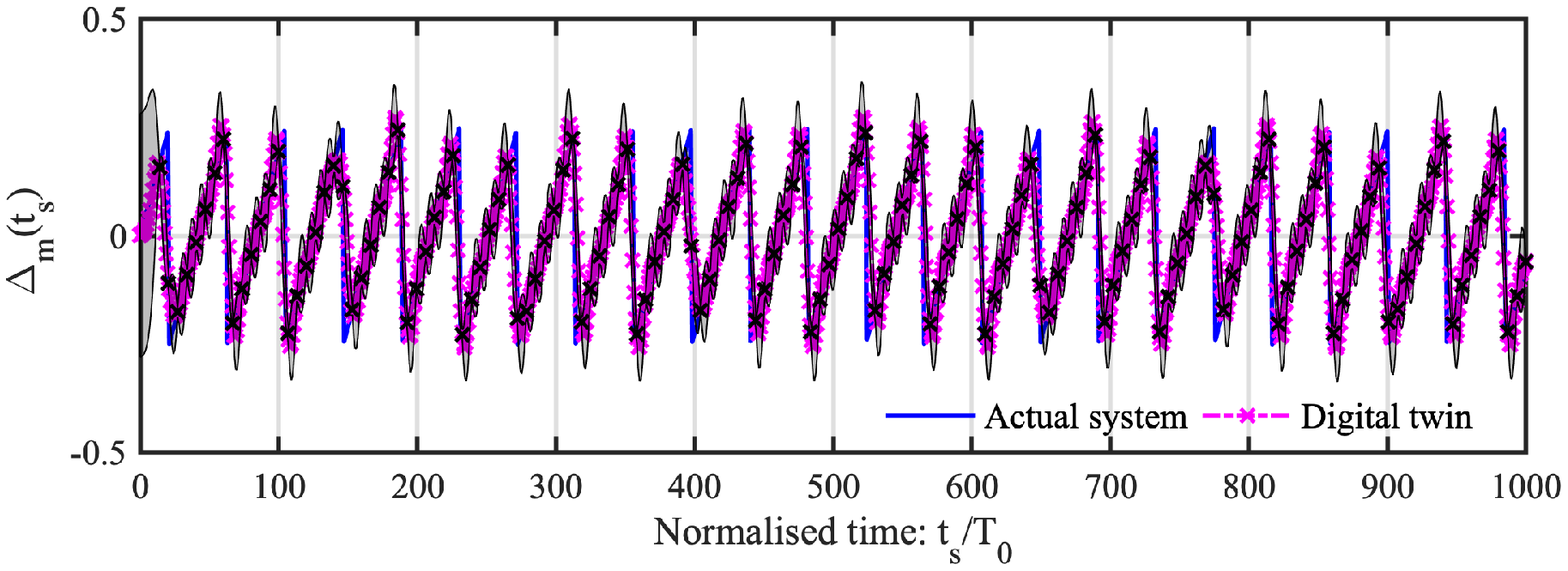}}
    \subfigure[$\sigma_{\theta} = 0.015$]{\includegraphics[width=0.8\textwidth]{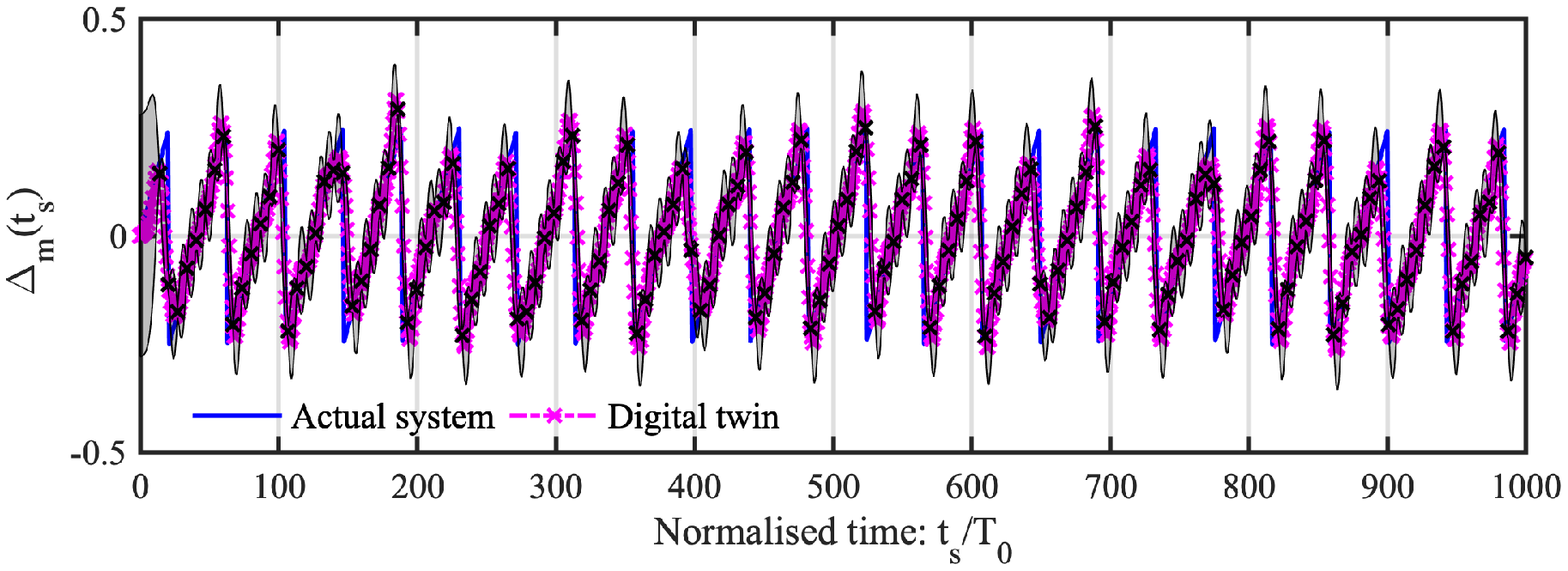}}
    \subfigure[$\sigma_{\theta} = 0.025$]{\includegraphics[width=0.8\textwidth]{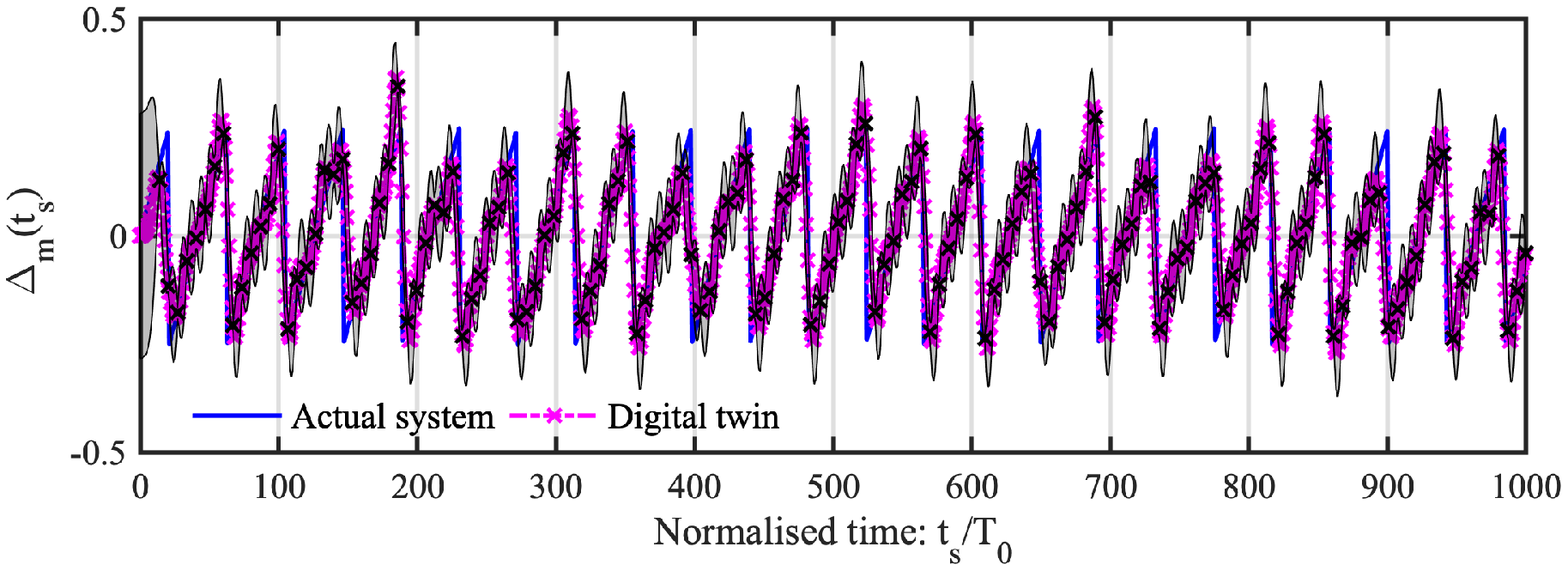}}
    \caption{Mass evolution as a function of the normalised slow time $t_s / T_0$ for GP based digital twin (simultaneous mass and stiffness evolution) obtained from 150 noisy data. Noise levels of $\sigma_{\theta} = 0.005$, $\sigma_{\theta} = 0.015$ and $\sigma_{\theta} = 0.025$ are considered. Bayesian information criteria yields `Linear' basis and `ARD Matern 5/2' covariance kernel. The shaded plot depicts the 95\% confidence interval.}
    \label{fig:gp_dt_mass_stiff_evol_noisy2}
\end{figure}

\begin{figure}[ht!]
    \centering
    \includegraphics[width=0.8\textwidth]{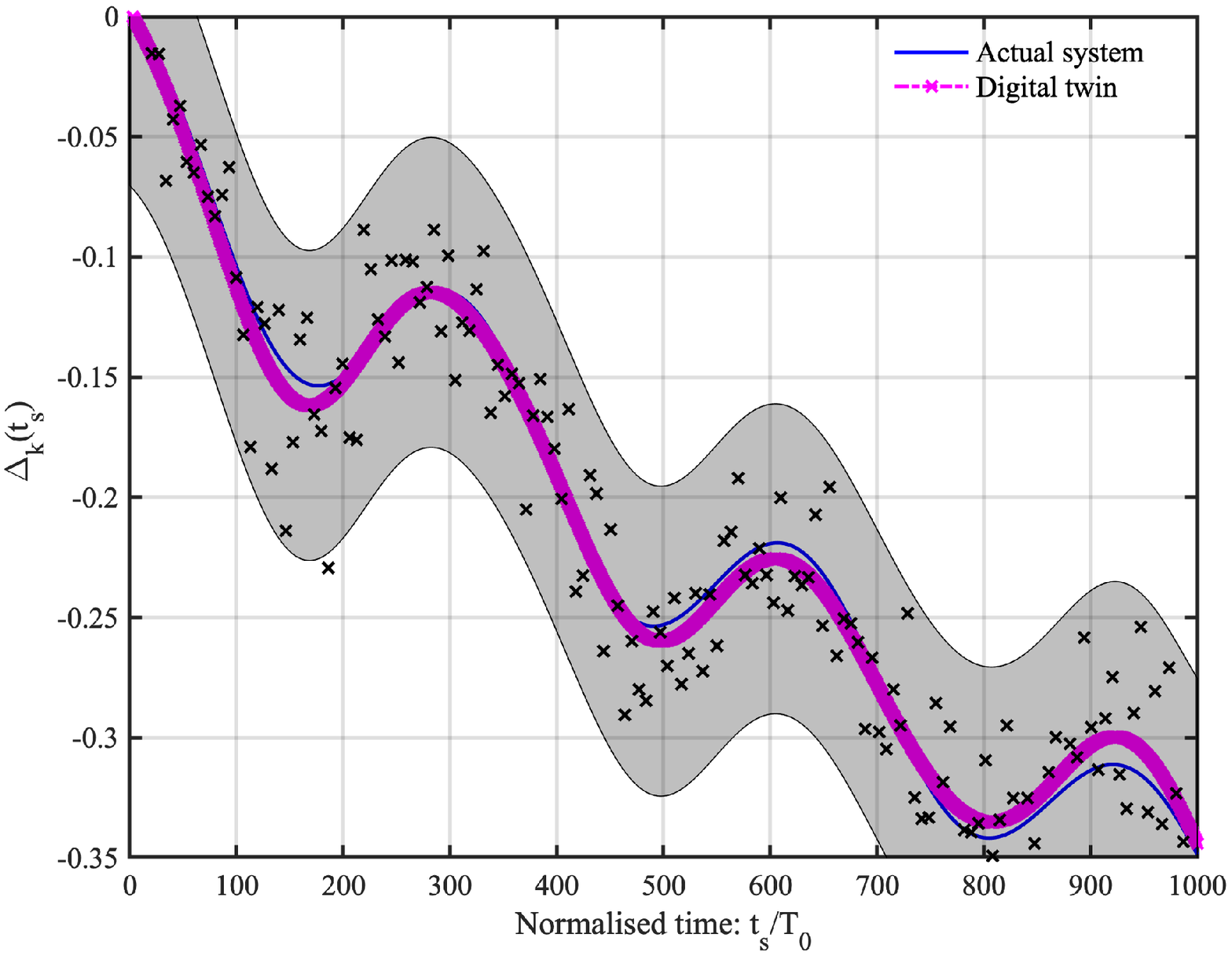}
    \caption{Stiffness evolution as a function of the normalised slow time $t_s / T_0$ for GP based digital twin (simultaneous mass and stiffness evolution) trained with 150 noisy data with $\sigma_{\theta} = 0.025$. Bayesian information criteria yields `Linear' basis and `ARD Matern 5/2' covariance kernel. The shaded plot depicts the 95\% confidence interval.}
    \label{fig:gp_dt_mass_stiff_evol_noisy3}
\end{figure}

\begin{figure}[ht!]
    \centering
    \includegraphics[width=0.8\textwidth]{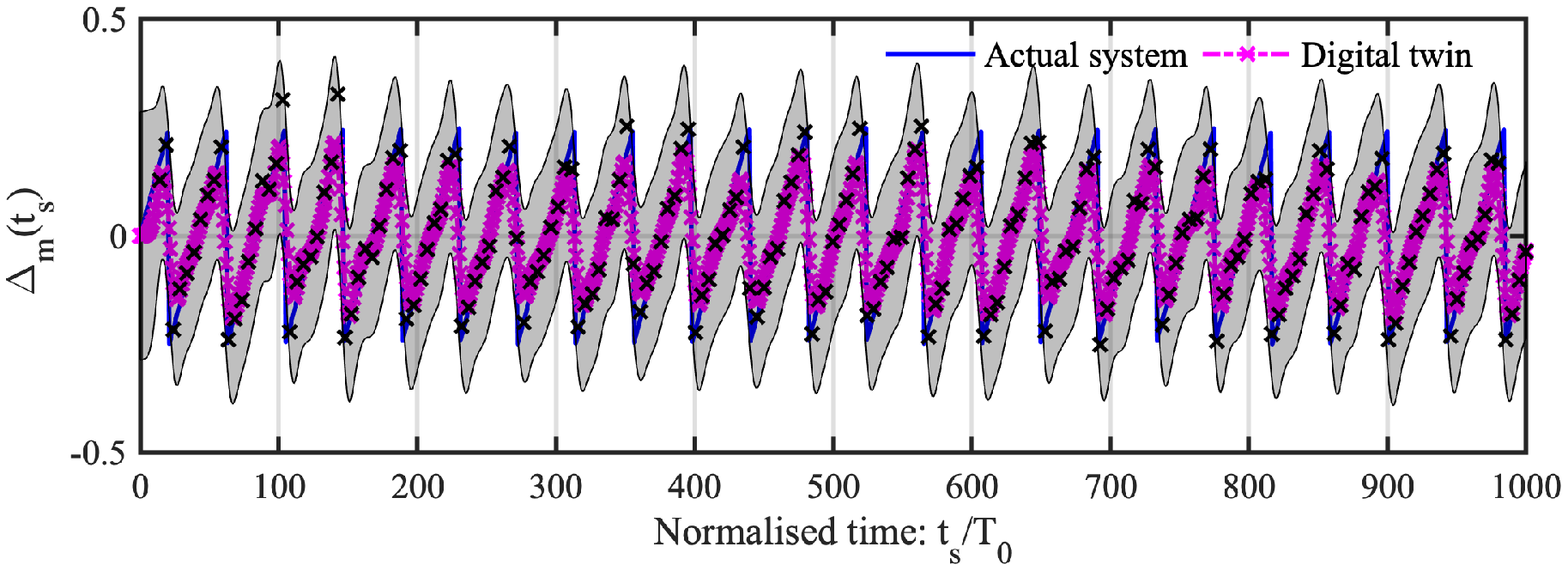}
    \caption{Mass evolution as a function of the normalised slow time $t_s / T_0$ for GP based digital twin (simultaneous mass and stiffness evolution) trained with 200 noisy data with $\sigma_{\theta} = 0.025$. Bayesian information criteria yields `Linear' basis and `ARD Matern 5/2' covariance kernel. The shaded plot depicts the 95\% confidence interval.}
    \label{fig:gp_dt_mass_stiff_evol_noisy4}
\end{figure}

Figs. \ref{fig:gp_dt_mass_stiff_evol_noisy2} and \ref{fig:gp_dt_mass_stiff_evol_noisy3} show the evolution of mass and stiffness of the GP based digital twin trainedwith 150 noisy observations. For mass evolution, results corresponding to all three noise levels are shown. With the increase in the number of observations, we observe a dramatic improvement in the GP assisted digital twin. The evolution of stiffness is shown only for $\sigma_{\theta} = 0.025$ and results obtained are highly accurate.Lastly, evolution of mass with 200 noisy observations and $\sigma_{\theta} = 0.025$ is shown in \autoref{fig:gp_dt_mass_stiff_evol_noisy4}. As expected, this yields the best results. The evolution of stiffness for 200 observations are found to be similar to those with 150 samples and hence, the same is not shown in this paper.

\section{Discussion}\label{sec:discussions}

Although in theory, a digital twin of a physical system can be achieved in several ways, most of the existing works on digital twin have focused on the broader conceptual aspect. In this paper, we take a different path and focus on a specific case of a structural dynamical system. More specifically, we have considered surrogate models such as GP with a single degree of freedom system. The key ideas proposed in this paper include:
\begin{itemize}

  	\item We introduce the concept of a surrogate model within the digital twin technology. In particular, we advocated the use of GP within the digital twin technology. The GP was trained based on observations on a slow time-scale and was used for predicting the model parameters in the fast time-scale.
	
	\item The importance of collecting more data during the system life-cycle is illustrated in this study. A dramatic improvement in the performance of the digital twin was observed with an increase in the number of data collected. In other words, over the same window of time, a higher sampling rate is needed. 
	
	\item The importance of cleaner data in the context of a digital twin is illustrated in this paper. With less noise in the data, the digital twin can quickly track the time evolution of the model parameters. Digital signal processing algorithms can be used for data cleaning \cite{roy2005helicopter, ganguli2002noise}.
	
	\item GP, being a Bayesian surrogate model, can quantify the uncertainty in the system due to limited data and noise. These uncertainties can be used for deciding if there is a need to increase the frequency of data collection, i.e. increase the measurement sampling rate.
	
	\item Overall, GP was able to capture both mass and stiffness variation from limited noisy data; although, for capturing the time evolution of mass, 	more observations are needed as compared to the stiffness evolution case. 
	
\end{itemize}

The system studied here is a simple single-degree-of-freedom dynamical system governed by second-order ordinary differential equations. The same framework, as it is, applies to other physical systems (e.g., simple electrical systems involving a resistor, capacitor and an inductor) governed by this kind of equations. The framework presented here can be extended for more rigorous investigations encompassing a wider variety of practical problems. Some of the future possibilities are highlighted below:
\begin{enumerate}

  \item {\it Surrogate assisted digital twin for big data: } In this paper, natural frequency measurements are used for establishing the GP based digital twin.  However, it will be more useful if the GP based digital twin can be trained directly from the time-history of measured responses.  This is essentially a big-data problem and {\it vanilla} GP is unable to tackle such problems.  More advanced versions of GP, such as the sparse GP \cite{snelson2006sparse} and convolution GP \cite{alvarez2011computationally} may be explored in that case.

  \item {\it Surrogate based digital twin for continuum system:} For many engineering problems, a continuum model represented by the partial differential equation is the preferred choice.  Developing a surrogate-based digital twin for such a system will be of great use.  This is primarily a sparse data scenario as sensor responses will only be available at few spatial locations.  Methods such as convolution neural networks \cite{kim2014convolutional} can be used for such systems.
   
  \item {\it Digital twin for systems with unknown/imperfect physics: } In literature, there exists a wide range of problems for which the governing physical law is not well defined.  It will be interesting and extremely useful to learn/develop the digital twin purely based on data. Works on discovering physics from data by using machine learning can be found in the literature \cite{xu2019dl}. Genetic programming can also be used to fit mathematical functions to data \cite{singh2007genetic}.     
   
  \item {\it Surrogate based digital twin models with machine learning: } Over the past few years, the machine learning community has witnessed rapid progress with developments of techniques such as deep neural networks \cite{goswami2020transfer,chakraborty2020simulation,chakraborty2020transfer}, convolution neural networks \cite{kim2014convolutional}, among others.  Using these techniques within the digital twin framework can possibly push this technology to new heights.  The required sampling rate and data quality could be reduced by using machine learning methods.
   
  \item {\it Multiple-degrees-of-freedom (MDOF) digital twins using surrogate models: } In this paper,  we focused on a single-degree-of-freedom (SDOF) model, which can be considered to be a simple idealisation of complex multiple-degrees-of-freedom (MDOF) systems.  However, for more effective digital twin with realistic predictive capabilities, MDOF digital twin should be considered. To that end, it is necessary to develop surrogate-based MDOF digital twins.
   
  \item {\it Surrogate models for nonlinear digital twins:}  The model considered in this paper is a linear ordinary differential equation.  However, many physical systems exhibit nonlinear behaviour.  A classical example of a nonlinear system is the Duffing oscillator \cite{Nayfeh93}.  For such a non-linear system,  it is unlikely that a closed-form solution will exist and hence,  developing a surrogate-based digital twin model will be immensely useful.

  \item {\it Forecasting using surrogate models for digital twins: }   One of the primary tasks of a digital twin is to provide a future prediction.  With the physics-based digital twin proposed previously \cite{Ganguli2020},  it is not possible to predict the future responses.  The GP based digital twin model, on the contrary, can predict the short term future response.  It is necessary to develop surrogate-based digital twin capable of forecasting the long-term response of the system.

  \item {\it Predicting extremely low-probability catastrophic events for digital twins with surrogate models:} Till date, application of digital twin technology is mostly limited to maintenance, prognosis and health monitoring.  Other possible applications of digital twin involve computing probability of failure, rare event probability and extreme events.

  \item {\it High-dimensional surrogate models for multi time-scale digital twins: }   In this paper, we assumed one time-scale for the evolution of the digital twin.   However, there is no physical or mathematical reason as to why this must be restricted to only one time scale.   It is possible that the parameters may evolve at different time-scales.  Such systems are known as multi-scale dynamical system \cite{Chakraborty2018efficient}.   Developing surrogate-based digital twin for such multi-scale dynamical systems needs further investigation.
   
  \item {\it Hybrid surrogate models for digital twins: } In this paper only one type of surrogate model, namely, the Gaussian process emulator was used. Several types of surrogate models have been employed to solve a wide range of complex problems with different numbers of variables \cite{Chakraborty2019graph,singh2007genetic}. It is well known that some surrogate models perform better than the others in certain situations. It is therefore perfectly plausible to employ multiple surrogate models simultaneously by taking the benefit of their relative strengths. Such hybrid surrogate digital twins are expected to perform in a superior manner compared to single surrogate digital twins.

\end{enumerate}
\section{Conclusions}\label{sec:conclusions}

Surrogate models have been developed over the past four decades with increasing computational efficiency, sophistication, variety, depth and breadth. They are a class of machine learning methods which thrive on the availability of data and superior computing power. As digital twins also expected to exploit data and computational methods, there is a compelling case for the use of surrogate models in this context. Motivated by this synergy, we have explored the possibility of using a particular surrogate model, namely, the Gaussian process (GP) emulator, for the digital twin of a damped single-degree-of-freedom dynamic system. The proposed digital twin evolves at a time-scale which is much slower than the dynamics of the system. This makes it possible to identify crucial system parameters as a function of the `slow-time' from continuously measured data. Closed-form expressions derived considering the dynamics of the system in the fast timescale have been employed. The Gaussian process emulator is employed in the slow timescale, where the impact of lack of data (sparse data) and noise in the data have been explored in detail.

The results arising from applying the Gaussian process emulator show that sparsity of data may prevent the GP from accurately capturing the evolution of mass and stiffness. This is particularly true for the mass evolution case where a higher sampling rate is necessary for accurately capturing the time evolution of mass. It is to be noted that the uncertainty in the GP based digital twin model results from both sparsity in the data and noise in the measurements. Consequently, only increasing the number of observations may not necessarily reduce the uncertainty in the system. On the other hand, collecting clean data (i.e., data with lower noise level) helps GP with tracking the evolution of mass and stiffness evolution more accurately. Nonetheless, even for the cases with a relatively larger noise level, GP is found to yield an accurate result. Moreover, GP also captures the uncertainty due to limited and sparse data.  

Although only the Gaussian process emulator with single-degree-of-freedom dynamic systems are considered, several conceptual extensions directly followed from this work are described in detail. The approach proposed here exploits closed-form expressions based on the physics of the system and a surrogate model for the measured data. 
However, the proposed surrogate model based approach is perfectly suitable when more advanced computational simulation to be used for the physical problem (for example, when multi-degree-of-freedom system is considered). Therefore, the underlying formulation is not restricted to the single-degree-of-freedom example used here.
The overall framework provides a paradigm for a hybrid physics-based and data-driven digital twin approach. The physics-based approach is related to the `fast time', while the data-driven approach is operational on the `slow time'. The separation of scientific approaches based on the inherently different time-scales will allow the integration of multi-disciplinary approaches in the future development of the digital twin technology.  
\appendix
\section{Bayesian information criteria}
\label{app}
In this paper, the optimal order of basis function and the optimal covariance kernel for GP are determined by using the Bayesian information criteria.
The Bayesian information criteria corresponding to the $m$-th model is defined as \cite{Murphy2012}
\begin{equation}\label{eq:bic}
    bic_m = k_m\log (n) - \mathcal L\left(\bm \hat \theta_m \right),
\end{equation}
where $k_m$ is the number of parameters and $n$ represents the number data-point available to the $m$-th model. 
$\mathcal L (\bm \theta )$ in \autoref{eq:bic} represents the data-likelihood for the $m$-th model.
The first term in \autoref{eq:bic} penalizes a complex model whereas the second term ensures that the model that best explains the model is selected.
For GP, $\mathcal L (\bm \theta _m )$ follows a multi-variate Gaussian distribution \cite{williams2006gaussian}.

For determining the optimal model using the BIC criteria, models corresponding to all possible combinations from mean function and covariance function is considered.
The possible candidates for mean and covariance functions are shown in \autoref{tab:mean_cov_pool}.
There are three possible mean function and 10 possible covariance functions. Therefore, total 30 models were considered in this study.
The BIC score is evaluated for each of the 30 models and the one with the minimum BIC score is selected. The steps involve are shown in \autoref{alg:bic_based_selection}.
\begin{algorithm}
\caption{Model selection using Bayesian information criteria}\label{alg:bic_based_selection}
\textbf{Initialize: }Provide the pool of mean function and covariance function. Also provide the training data-set, $\mathcal D$.\\
Formulate all possible models ($M$) by using the pool of mean function and covariance function. \\
\For{$m = 1,\ldots,M$}{
$\bm {\hat \theta}_m \leftarrow \arg\max_{\bm \theta} \mathcal L (\bm \theta )$ \Comment*[r]{Train GP}
Use $\hat {\bm \theta}_m$ to compute $bic[m]$ \Comment*[r]{\autoref{eq:bic}}  
}
Select the model with the lowest $bic$ score.
\end{algorithm}

\section*{Acknowledgements}

RG acknowledges the financial support from The Royal Academy of Engineering through the award of a Distinguished Visiting Fellowship, grant number DVFS21819/9/5. SA  acknowledges the financial support from The Engineering Physical Science Research Council (EPSRC) through a programme grant EP/R006768/1.



\end{document}